\documentclass{article}

\usepackage{arxiv}

\usepackage[utf8]{inputenc} 
\usepackage[T1]{fontenc}    
\usepackage{hyperref}       
\usepackage{url}            
\usepackage{booktabs}       
\usepackage{amsfonts}       
\usepackage{nicefrac}       
\usepackage{microtype}      
\usepackage{cleveref}       
\usepackage{graphicx}
\usepackage{natbib}
\usepackage{doi}
\usepackage{xcolor}

\title{Predicting household socioeconomic position in Mozambique using satellite and household imagery}

\usepackage{authblk}

\newbox{\orcid}\sbox{\orcid}{\includegraphics[scale=0.06]{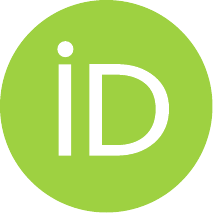}} 
\author[1,2]{%
	\href{https://orcid.org/0000-0003-0470-0760}{\usebox{\orcid}\hspace{1mm}Carles Milà\thanks{\texttt{carles.mila@isglobal.org}}}%
}
\author[3]{%
	{\hspace{1mm}Teodimiro Matsena}%
}
\author[3]{%
	{\hspace{1mm}Edgar Jamisse}%
}
\author[1,3]{%
	{\hspace{1mm}Jovito Nunes}%
}
\author[1,3,4,5,6,7]{%
	\href{https://orcid.org/0000-0003-0875-7596}{\usebox{\orcid}\hspace{1mm}Quique Bassat}%
}
\author[1]{%
	\href{https://orcid.org/0000-0002-5030-1915}{\usebox{\orcid}\hspace{1mm}Paula Petrone}%
}
\author[1,3,7,8]{%
	\href{https://orcid.org/0000-0002-2499-2732}{\usebox{\orcid}\hspace{1mm}Elisa Sicuri}%
}
\author[3,7]{%
	\href{https://orcid.org/0000-0002-8724-0319}{\usebox{\orcid}\hspace{1mm}Charfudin Sacoor}%
}
\author[1,2,3,4]{%
	\href{https://orcid.org/0000-0003-3919-8264}{\usebox{\orcid}\hspace{1mm}Cathryn Tonne}%
}
\affil[1]{ISGlobal, Barcelona, Spain}
\affil[2]{Universitat Pompeu Fabra (UPF), Barcelona, Spain}
\affil[3]{Centro de Investigação em Saúde de Manhiça (CISM), Maputo, Mozambique}
\affil[4]{CIBER epidemiología y salud pública (CIBERESP), Madrid, Spain}
\affil[5]{ICREA, Barcelona, Spain}
\affil[6]{Pediatric Department, Hospital Sant Joan de Deu - Universitat de Barcelona, Barcelona, Spain}
\affil[7]{Facultat de Medicina i Ciències de la Salut, Universitat de Barcelona (UB), Barcelona, Spain}
\affil[8]{LSE Health, Department of Health Policy, London School of Economics 
and Political Science, London, UK}

\renewcommand{\shorttitle}{Predicting household SEP using satellite and household imagery}

\hypersetup{
pdftitle={Predicting household socioeconomic position in Mozambique using satellite and household imagery},
pdfauthor={Carles Milà, Teodimiro Matsena, Edgar Jamisse, Jovito Nunes, Paula Petrone, Elisa Sicuri, Charfudin Sacoor, Cathryn Tonne}
}

\begin{document}
\maketitle

\begin{abstract}
Many studies have predicted SocioEconomic Position (SEP) for aggregated spatial units such as villages using satellite data, but SEP prediction at the household level and other sources of imagery have not been yet explored. We assembled a dataset of 975 households in a semi-rural district in southern Mozambique, consisting of self-reported asset, expenditure, and income SEP data, as well as multimodal imagery including satellite images and a ground-based photograph survey of 11 household elements. We fine-tuned a convolutional neural network to extract feature vectors from the images, which we then used in regression analyzes to model household SEP using different sets of image types. The best prediction performance was found when modeling asset-based SEP using random forest models with all image types, while the performance for expenditure- and income-based SEP was lower. Using SHAP, we observed clear differences between the images with the largest positive and negative effects, as well as identified the most relevant household elements in the predictions. Finally, we fitted an additional reduced model using only the identified relevant household elements, which had an only slightly lower performance compared to models using all images. Our results show how ground-based household photographs allow to zoom in from an area-level to an individual household prediction while minimizing the data collection effort by using explainable machine learning. The developed workflow can be potentially integrated into routine household surveys, where the collected household imagery could be used for other purposes, such as refined asset characterization and environmental exposure assessment.
\end{abstract}

\newpage
\section{Introduction}

SocioEconomic Position (SEP), defined as the set of social and economic factors that determine the position of an individual or a group within a society, is an important, complex, and broadly used concept in health research \citep{Galobardes2006}. SEP is an important health determinant across the life course through multiple pathways such as the access to healthcare, health-related behaviours, and environmental and occupational exposures that may lead to health inequalities \citep{Adler2010}. Furthermore, it is also one of the strongest confounders in environmental epidemiology \cite[e.g.,][]{Hystad2019} due to a potential relationship between SEP and many exposures. 

Given its multidimensional nature, there are many ways to measure SEP and the appropriate measure depends on the research question and context \citep{Galobardes2006}. In Low- and Middle-Income Countries (LMICs), one of the most used approaches are asset-based measures, which are widely available through Demographic and Health Surveys (DHS) \citep{Filmer2001}. Asset-based SEP is constructed based on information on durable assets, housing materials, and access to basic services of a given household, which are then summarized into a single indicator using statistical methods \citep{filmer2012}. While collecting asset information is easier than other SEP metrics, asset-based SEP is a relative measure that may lack variability if the set of variables to be included is not carefully chosen \citep{Howe2012} depending on the type of area (urban vs. rural) and geography (variation between and within countries) \citep{Howland2021}. While income-based SEP is extensively used in high-income countries \citep{Galobardes2006}, in LMICs income can be more difficult to measure due to a higher prevalence of informal and seasonal labour, multiple income sources, home production, and income in the form of goods \citep{Howe2012}. Expenditure SEP measures have also been used: although these are more stable over time than income, they are challenging to measure due to misreporting, limitations of expenditure diaries, the monetization of home-produced goods \citep{Howe2012}, and seasonal fluctuations in food prices \citep{Bai2020}. 

Collecting SEP data is complex, time-consuming, and costly \citep{Burke2021}. Motivated by this, a rapidly growing body of literature focusing on predictive mapping approaches to estimate SEP based on remote sensing data has emerged, leveraging the increasing availability of satellite data \citep{Burke2021} and advances in the fields of machine and deep learning \citep{Gao2019}. Many of the studies focused on the prediction of asset-based SEP, while others predicted expenditure, income, education, or other poverty-related measures \citep{Hall2023}. Early works focused on SEP prediction based on nighttime light satellite data \cite[e.g.,][]{Elvidge2009}, but recently works using very high-resolution (<10m) optical remote sensing data have become increasingly prevalent in LMICs \cite[e.g.,][]{head2017} including Mozambique \citep{Sohnesen2022}. Alternatively, freely available high- and medium-resolution optical satellite data (MODIS, Landsat, Sentinel 2) have also been used \cite[e.g.,][]{Yeh2020, Hersh2021}. Many of these studies relied on deep Convolutional Neural Networks (CNN) to extract feature vectors from images \cite[e.g.,][]{Jean2016, zhao2020}, but some studies also used hand-crafted features such as vehicle, building, or roof object detection \cite[e.g.,][]{Ayush2020, engstrom2022}; vegetation and imperviousness spectral indices \cite[e.g.,][]{Niu2020}; or features derived from land cover classification and texture analyses \cite[e.g.,][]{Duque2015}. 

A recent review of SEP mapping studies found that the factors that benefited performance included the combination of deep learning and machine learning methods (as opposed to using only one of them), predicting asset-based measures (compared to expenditure and income), and the number of datasets used \citep{Hall2023}. On the other hand, the most important limitation was the scarce and noisy ground truth data \citep{Burke2021}. Data scarcity has been addressed via transfer learning of deep learning models \citep{Burke2021}, with some studies using widely available SEP proxies such as satellite nighttime lights \cite[e.g.,][]{Jean2016, xie2016}.

Despite rapid progress in the field of SEP prediction, there are still some challenges to be addressed. First, most mapping studies focused on predicting SEP for aggregated spatial units (e.g., villages \citep{Yeh2020}, grid cells \citep{chi2022}), while studies that predict SEP at the household level are still infrequent \citep{watmough2019}. Targeting studies that estimate SEP for individual households are important \citep{mcbride2022} since they allow to better allocate resources to specific households during interventions (e.g., food or energy-related), address inequalities within communities, and enable the use of predictions in epidemiological analyses. Second, the use of ground imagery such as street view images could complement satellite imagery and increase model performance. However, studies using street view imagery have been mostly limited to urban areas in high-income countries \cite[e.g.,][]{Gebru2017, suel2019, Fan2023}  due to data availability. In addition to street view images, household imagery such as the Dollar Street dataset \citep{rojas2022} would be particularly useful to go beyond simple binary asset indicators and capture information about their quality \citep{Howe2012}. Third,  there is a need to increase the interpretability and explainability of SEP predictive models \citep{hall2022}. As examples, SHAP values (SHapley Additive exPlanations) have been used to identify the most relevant features in SEP models \citep{Ayush2020}, whereas Grad-CAM has been used to compute activation maps in satellite images used to predict SEP \citep{abitbol2020}. Explainable machine learning methods are especially important in multimodal datasets \cite[e.g.,][]{chi2022}, where identifying relevant and redundant modalities is key to finding opportunities to optimize data collection.

This study aims to advance the current literature on SEP prediction by introducing: 1) a multimodal dataset that combines very high-resolution satellite imagery and a household photograph survey, 2) a SEP prediction at the household level, 3) a comprehensive evaluation of SEP by using multiple measures, and 4) the use of explainable machine learning to improve model explainability and optimize sampling. The objectives of the study are the following:
\begin{itemize}
    \item To predict household SEP using a multimodal satellite and household imagery dataset.
    \item To assess the contribution of each of the images to the SEP prediction to explain the models.
    \item To optimize future data collection efforts by identifying the most relevant image types in the models.
\end{itemize}

\section{Materials and methods}

\subsection{Study area and population} \label{ref:studypop}

We used data from the project "Novel methods to assess household socioeconomic position in Manhiça district in Mozambique", co-led by the \textit{Centro de Investigação em Saúde de Manhiça} (CISM) and the Barcelona Institute for Global Health. The study area of the project was the semi-rural \textit{Manhiça sede} administrative unit located within the Manhiça district in southern Mozambique. \textit{Manhiça sede} covers a surface of 412 km$^2$ (supplementary Figure S1) and has a population of 78,479 inhabitants with a median household size of six persons \citep{Nhacolo2021}. The study area includes a relatively dense populated area close to the main road, while the rest of the study area is sparsely populated (supplementary Figure S1). The study area is covered by the Health and Demographic Surveillance System (HDSS) managed by CISM, which routinely collects demographic and economic information from all households \citep{Nhacolo2021}. 

\subsection{Sampling procedure}

The target sample size was 1,000 households, for which stratified sampling was used to ensure a comprehensive representation of SEP. To do so, we first compiled data for all households residing in the study area from the last available HDSS round at the time of the start of the study. With these data, we constructed an asset-based SEP measure (with the same methods as in the main analysis, see section \ref{ref:construction}) and categorized it into quartiles. Finally, we randomly selected 250 households per quartile for sampling, as well as some additional households as backup for cases where the home was deserted or the head of the household was repeatedly unavailable or refused to participate in the study. 

\subsection{Data collection} \label{ref:methods_data}

\begin{figure}[!th]
\centering 
\includegraphics[width=0.6\textwidth]{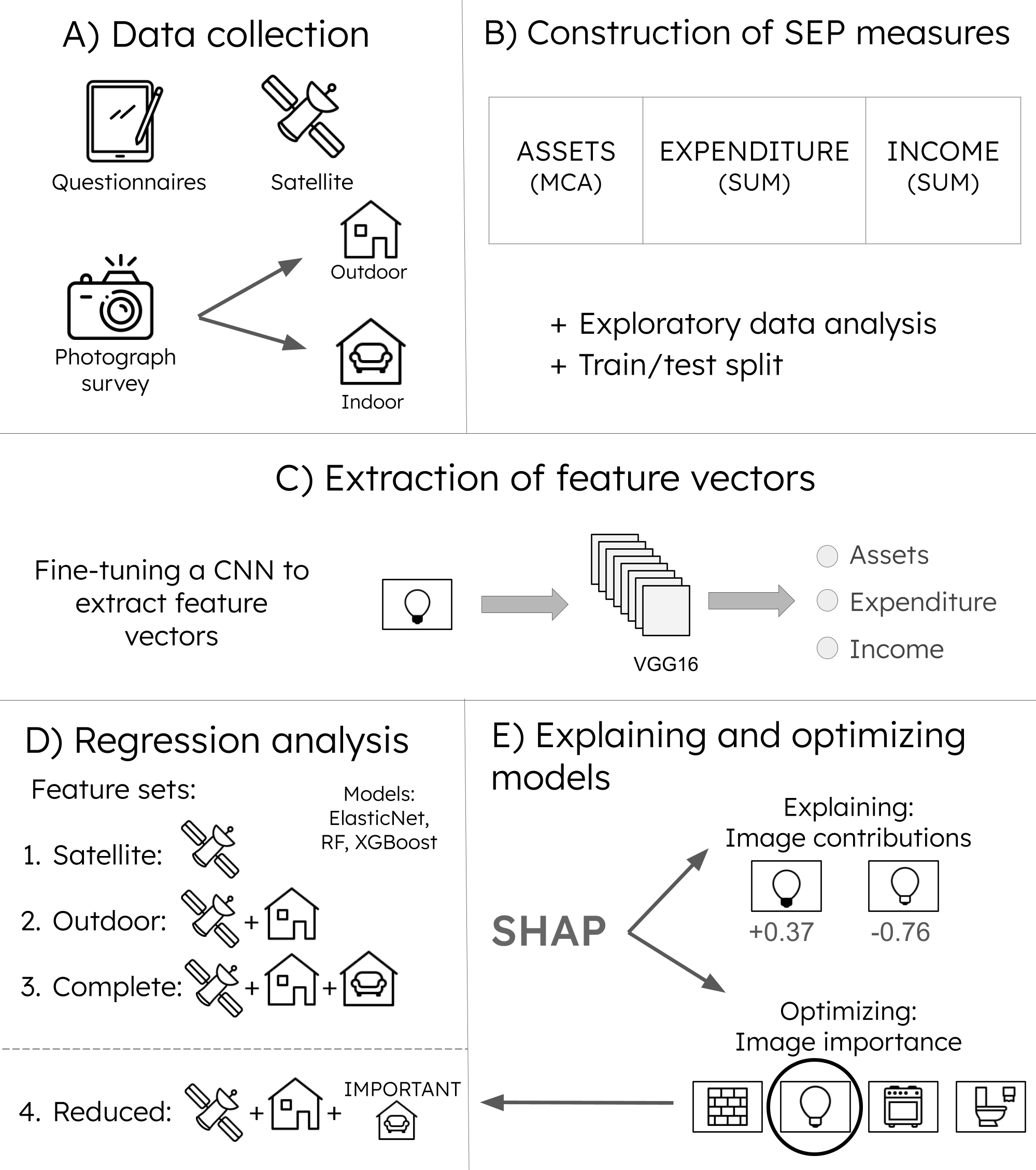}
\caption{Workflow of the data collection and analysis in the SEP study.}
\label{fig:workflow}
\end{figure}

Data collection for the SEP study (Figure \ref{fig:workflow}A) involved three different data sources: primary data collection of questionnaire and photograph data, secondary data collection from existing HDSS data, and satellite data.

\subsubsection{Primary data collection}

Primary data collection visits to sampled households were structured in three phases. First, a field worker explained the goals and scope of the study to the head of the household, who was invited to participate. If the response was positive, the participant was asked to sign an informed consent form that contained all the information regarding the study, the data to be collected, how these data would be analyzed and stored, and the participants' rights, which included the possibility not to respond to any of the questions, as well as refusing any of the photographs to be taken. The participant also had the right to withdraw their consent at any time during or after data collection. 

Second, a questionnaire including a series of questions on monthly average income and expenditure sources (available in supplementary Table S1) was verbally administered. The questionnaire was adapted from the Living Standards Measurement Study templates from the World Bank \citep{oseni2021}. Finally, a photograph survey was conducted in which the field worker photographed 11 elements of the household following a standardized protocol. The household elements comprised outdoor elements that could always be photographed from the exterior of the home: front door, wall, street view; and indoor elements, which could be located in the interior of the houses: roof, floor, light source, kitchen, stove, bathroom, latrine, and water source (see examples in Figure \ref{fig:photos}).

All fieldworkers used the same model of tablet device with a built-in camera for both questionnaire and photograph data collection (Samsung Galaxy Tab A (2018, LTE) SM-T595 32GB 10.5". Primary Camera: 8 MP, AF, f/1.9, LED flash, panorama, 1080p@30fps). During the visit, the coordinates of the front door of the main building of the household were registered using the same device. Data collection took place between July and November 2022 following a 30-participant pilot designed to identify and address potential issues. Continuous quality control of the data was established to identify photographs that did not follow the protocol. In such cases, the non-compliant photograph was repeated whenever possible.

\subsubsection{Secondary data collection}

We also used existing asset data from the Manhiça HDSS for the sampled households. We extracted data from the HDSS update round that occurred in parallel to our data collection to ensure temporal overlap between the two data sources. Asset data included a series of questions on self-reported possessions and building materials (listed in supplementary Table S2) that have previously been used to measure SEP in the area \citep{grau2022}. 

\subsubsection{Satellite imagery}

\begin{figure}[!th]
\centering 
\includegraphics[width=0.95\textwidth]{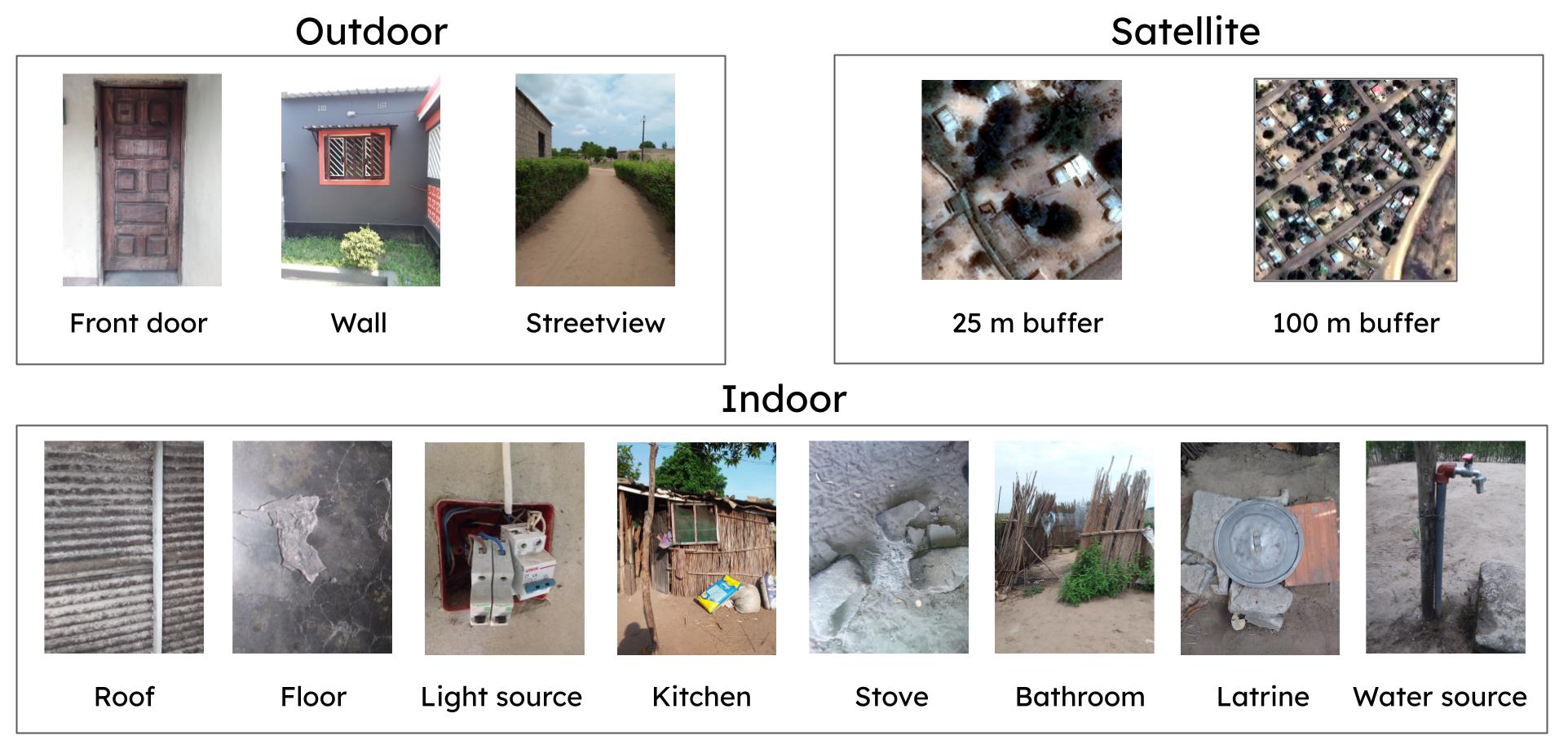}
\caption{Examples of all image types collected in the study.}
\label{fig:photos}
\end{figure}

We obtained very high-resolution satellite images for the study area (Worldview 2, 30 cm resolution pan-sharpened, DigitalGlobe, Inc. (2021), provided by European Space Imaging) through the European Space Agency (ESA) Third Party Missions scheme. The acquisition date of the images was the 21\textsuperscript{st} of October 2021. As pre-processing, we restricted the range of the red, green, and blue channels to percentiles 1-99\% of the pixel values to remove artefacts, which were mostly related to sun glitter on metal roofs. Then, we obtained two satellite images per household by extracting the area surrounding the geocoded residences. To do so, we defined 25 m (dwelling and immediate surroundings) and 100 m square buffers (dwelling and surrounding neighbourhood) around each of the household geocodes, and cropped the satellite images to the computed extent (Figure \ref{fig:photos}). 

\subsubsection{Final dataset and missing data} \label{sec:missing}

At the end of data collection, we obtained a dataset containing information on 975 households. No expenditure or income data were missing, while information on 24 assets (<0.1\% of the total) was not available. The percentage of missing photographs was lower than 3\% for all image types except for the kitchen category (supplementary Table S3). This was due to the large variety of kitchen settings in the study area (e.g., indoors vs. outdoors, uncovered vs. covered, with vs. without walls), which was clarified in the protocol during the first weeks of data collection.

\subsection{Analysis}

\subsubsection{Construction of SEP measures, exploratory analyses, and data split} \label{ref:construction}

We used the collected questionnaire data to construct three continuous SEP measures: income, expenditure, and assets (Figure \ref{fig:workflow}B). The computed SEP measures were considered the ground truth in the study. Income-based and expenditure-based SEP were created by summing the different sources of monthly average income and expenditure, respectively (listed in supplementary Table S1). We used aggregate household expenditure and income rather than per capita (with or without adjustment for scale economies) due to the increased correlation between assets and total expenditure \citep{Filmer2001}, which increases comparability between SEP measures \citep{poirier2024}. 

Asset-based SEP was constructed by first imputing missing assets with the most prevalent category in the data, and then using a Multiple Correspondence Analysis (MCA) with variables included in supplementary Table S2. Briefly, MCA is a dimensionality reduction method suitable for categorical data that has been used to construct asset-based SEP measures \citep{Traissac2012}, including studies in Manhiça \citep{grau2022}. The first MCA dimension was taken as the asset-based SEP measure, and we examined the captured inertia and column principal coordinates to interpret its composition. Note that this exact procedure was also used to create asset-based SEP used for sampling in section \ref{ref:studypop}. 

With the three created SEP measures, we performed an exploratory data analysis to inspect their distribution and correlation. Furthermore, we explored whether SEP differences in the photographs could be discerned by a human prior to any machine intervention. To do so, we randomly sampled, for each image type, one image from households within each SEP quartile. Finally, we performed a train/test data split by randomly assigning 800 and 175 households to each set and verified that the distributions of the three SEP measures in the two datasets overlapped.

\subsubsection{Extraction of feature vectors from the images} \label{ref:methods_extract}

We extracted feature vectors from all images by training a CNN aiming at predicting the three SEP measures and extracting the output values of the next to last fully connected layer (Figure \ref{fig:workflow}C). To do so, we used transfer learning with fine-tuning of an existing CNN. Transfer learning is a widely used strategy in SEP mapping studies (e.g., \citet{suel2019}) where the target dataset is small and has different characteristics from the source dataset \citep{elgendy2020}. We chose the VGG16 model trained on ImageNet \citep{simonyan2014} since VGG architectures have been successfully used for SEP prediction in previous studies (e.g.,\citet{head2017}). We adapted the VGG16 network to a multi-output classification aiming at predicting predict binary (above/below median) SEP according to the three SEP measures. We opted for a binary classification rather than a regression due to its easier nature considering our limited sample size.

\begin{figure}[!ht]
\centering 
\includegraphics[width=0.95\textwidth]{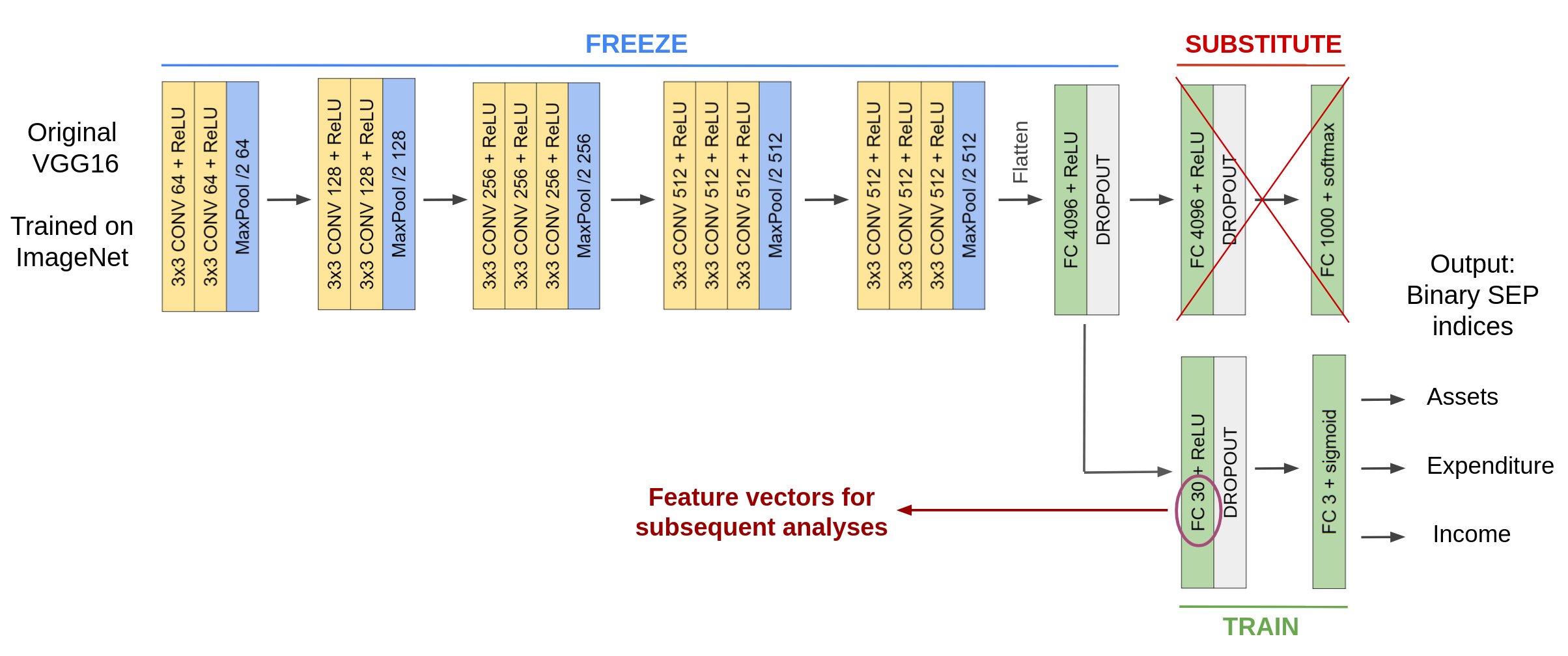}
\caption{Graphical representation of the transformed VGG16 CNN architecture used in the study for feature vector extraction.}
\label{fig:architecture}
\end{figure}

For each of the 13 image types, we downloaded weights for the VGG16 model and froze all layers. Then, we modified the output layer, which we changed to three nodes (one for each binary SEP measure) and set a sigmoid activation function for binary classification. We also modified the next to last fully connected layer, which we set to 30 nodes to make it more parsimonious. We enabled weight updating in the modified layers during backpropagation. A graphical description of the modified network is available in Figure \ref{fig:architecture}. We set the loss function to multi-label binary cross-entropy and used mini-batch gradient descent with momentum as the optimizer. We chose the model hyperparameters according to a set of experiments described in supplementary Methods S1. 

After defining the network and hyperparameters, we trained the CNN for each image type separately (i.e., 13 models were trained) using the 800 households in the training set. Models were trained for 50 epochs and data augmentation techniques were used. At the end of the training, we computed the percentage accuracy when predicting binary SEP in the test data. Finally, we extracted the output of dimension 30 of the next to last fully connected layer for each model, which we used as feature vectors in all subsequent analyses. Hence, a total of 30 features $\times$ 13 image types = 390 feature vectors were extracted for each household (provided no images were missing).

As an alternative to fine-tuning, we performed an alternative extraction of feature vectors using the output of the next to last fully connected layer of the original VGG16 network with ImageNet weights without any modification (i.e., 4,096 feature vectors per image type). The rationale behind this analysis was to assess the value of using a transfer learning approach vs. using off-the-shelf features from a pre-trained model \citep{Razavian2014}.

\subsubsection{Regression analyses} \label{ref:methods_prediction}

We used a supervised learning workflow to predict each of the three SEP measures, i.e. assets, expenditure, and income, using the feature vectors extracted from the images (Figure \ref{fig:workflow}D). The three outcomes were modelled as continuous variables and therefore were treated as regression problems. We defined three different predictors sets that went from the least to the most data collection effort to investigate to what extent they translated into a higher prediction accuracy:

\begin{itemize}
    \item Satellite: it only included satellite images.
    \item Outdoor: it included satellite images as well as household outdoor photographs.
    \item Complete: it included all image types including household indoor and outdoor photographs.
\end{itemize}

For each combination of predictor set and SEP measure, we then applied the following workflow: 
\begin{enumerate}
    \item We merged the SEP measure (outcome) and the feature vectors (i.e., 30 per image type) corresponding to the predictor set into two tables, one for the training data and another for the test data. 
    \item We defined a pipeline consisting of a simple mean imputation to account for missing photographs (see section \ref{sec:missing} and supplementary Table S3) followed by a regression modelling algorithm. As regression algorithms, we selected ElasticNet, a regularized linear model with L1 and L2 penalties \citep{hou2005}; Random Forest, a tree ensemble that uses bootstrap sampling with random feature selection \citep{breiman2001}; and XGBoost, a gradient boosting tree-based algorithm with regularization \citep{xgboostmod}. We tuned the model hyperparameters (listed in supplementary Table S4) using a randomized search based on 10-fold cross-validation. 
    \item We computed the prediction accuracy of the models by computing the Root Mean Square Error (RMSE) and the Pearson and Spearman correlation coefficients between the observed and predicted SEP values in the test data, as well as by graphically via scatterplots.
\end{enumerate}

In total, we fitted 3 (SEP measures) $\times$ 3 (predictor sets) $\times$ 3 (regression algorithms) = 27 models that were used to predict the test data, without counting the fits performed during cross-validation. 

As an alternative, we repeated the already-exposed supervised learning workflow but, rather than using the feature vectors extracted from the fine-tuned CNN, we used the off-the-shelf feature vectors directly extracted from the original VGG16 model as explained in section \ref{ref:methods_extract}. Since the number of features in this case was much larger than the number of observations (4,096 features per image type), we added a feature selection step prior to modelling that selected the $k$ top predictors according to F-statistics computed via univariate linear regression models. Here, $k$ was treated as a hyperparameter that was also tuned during cross-validation (supplementary Table S4).

\subsubsection{Explaining and optimizing models}

We performed a SHAP analysis of the most accurate complete regression model for each SEP measure (Figure \ref{fig:workflow}E). Briefly, SHAP is a model-agnostic explainable ML method based on coalitional game theory that estimates the linear additive effect of features when predicting an observation \citep{molnar2020}. SHAP values are expressed in outcome units and can be used as both a local or global explainability method \citep{lundberg2020}. We computed SHAP values in the train and test data by means of treeSHAP algorithm, an efficient method for tree-based models \citep{lundberg2020}. Next, we summed SHAP values from features derived from the same image (i.e., 13 SHAP values, one per image type, were available for each household after the sum).

The objective of the SHAP analysis was two-fold (Figure \ref{fig:workflow}E): First, we wanted to estimate the effect that each individual image had on the predicted SEP measures to explain the models. To do so, we selected the five images with the largest positive and negative SHAP values and visualized them. Secondly, we wanted to identify which image types were the most relevant for SEP prediction to optimize models. To do so, we examined the distributions of SHAP absolute values by image type, and ranked them according to their median. We identified the most relevant indoor household element as the image type with the largest median absolute SHAP value.

As a last step, we fitted "reduced" models using the "outdoor" predictor set as well as the identified most  important indoor image type for each SEP measure. To do so, we followed the same workflow described in section \ref{ref:methods_prediction} using the new predictor set (Figures \ref{fig:workflow}D and E). 

\subsection{Implementation and code availability}

We used R version 4.2 \citep{R} to run the MCA. The rest of analyses were performed in python version 3.11.7 \citep{python3}. The code to perform the data analysis and produce the figures and tables included in the article is available at \url{https://github.com/carlesmila/SEP-Mozambique}, where the packages used are also listed.

\subsection{Ethical considerations}

The SEP project received ethical approval by the bioethics institutional committee at CISM (protocol ID: CISM-015-2021), as well as by the Research Ethic Committee of Parc de Salut Mar in Spain (reference: 2022/10499). We gave special importance to the privacy of the participants by 1) anonymizing all data, 2) invalidating photographs with people in them, 3) drafting and following protocols regarding data access, storage, transfer, and security in the two participating research institutions, and 4) restricting the images included in the publication. Furthermore, we focused our analysis on the identification of relevant household elements so that the number of photographs that needs to be collected can be minimized in future studies, thus substantially reducing the burden placed on the participant.

\section{Results}

\subsection{Construction of SEP measures, exploratory analyses, and data split}

\begin{figure}[!th]
\centering 
\includegraphics[width=0.65\textwidth]{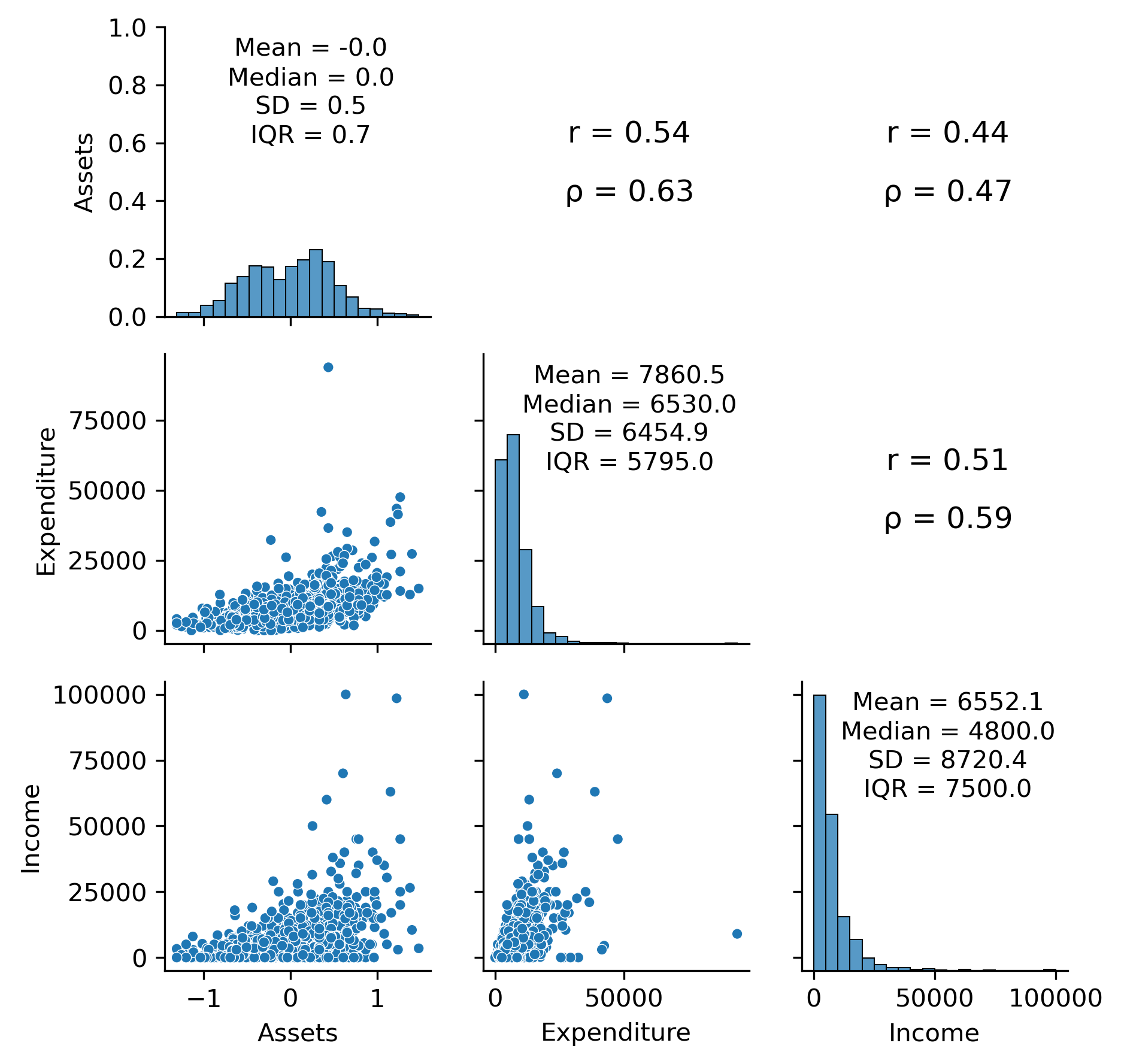}
\caption{Exploratory analysis of the SEP measures: assets (MCA first dimension), expenditure (metical), and income (metical). The diagonal panels show the histograms of the SEP measures and their descriptive statistics (SD: Standard Deviation; IQR: InterQuartile Range), while the lower and upper off-diagonal panels show the bivariate scatterplots and the Pearson ($r$) and Spearman ($\rho$) correlation coefficients, respectively.}
\label{fig:descriptive}
\end{figure}

The distribution of the three SEP measures was different (Figure \ref{fig:descriptive}): while income and expenditure-based SEP were right-skewed and had long tails, the distribution of asset-based SEP was more symmetrical. Correlations between SEP measures were moderate, being the highest between assets and expenditure, and the lowest between assets and income. When generating asset-based SEP using MCA, the resulting first dimension captured more than 80\% of the total inertia and showed a clear socioeconomic gradient (supplementary Figure S2): while large values in the first dimension were associated with high SEP assets (e.g., improved kitchen fuels and floors and piped water), the opposite was observed for low values (e.g., non-paved floors and absence of latrine). The distributions of the SEP measures in the train vs. test data overlapped (supplementary Figure S3).

\unskip

Imagery exploratory analyses showed differences between households in different SEP quartiles (supplementary Figure S4). For example, in the 'water source' category, photographs of the two first quartiles showed containers used to transport water from external sources to the residential compound, whereas upper quartiles showed piped water.

\subsection{Extraction of feature vectors from the images}

The fine-tuned VGG16 models were able to learn features associated with SEP. For all image types, the trained models were able to predict binary SEP measures with different ability in the test data (supplementary Table S5), with the best accuracy observed in light source models (0.77, 0.7 and 0.64 for asset, expenditure, and income-based SEP respectively), followed by latrine and kitchen models. Accuracies for satellite (100m) and street view models were the lowest.

\subsection{Regression analyses}

While the prediction accuracy for asset-based SEP in complete regression models was high, the performance of expenditure and income models was lower (Table \ref{tab:mainres}). Comparing the different sets of predictors, using all image types greatly benefited model performance for all SEP measures. Results for outdoor models fell in the middle of satellite and complete models, but still showed relatively good results for asset-based SEP. Differences between models using the same predictor set but different regression algorithms were modest, with random forest being the algorithm that performed the best in virtually all cases. Scatterplots of observed vs. predicted values in the test data reflected the gains in performance when using all images in complete models (supplementary Figures S5-S7).

\begin{table}[!ht]
\caption{Prediction accuracy measured with Pearson ($r$) and Spearman ($\rho$) correlation coefficients and RMSE in the test data by SEP measure, model type, and algorithm.}
\label{tab:mainres}
\centering
\begin{tabular}{|l|ccc|ccc|ccc|}
\hline
Model &  & Assets & & & Expenditure & & & Income & \\
 & $r$ & $\rho$ & RMSE & $r$ & $\rho$ & RMSE & $r$ & $\rho$ & RMSE \\
\hline
Satellite: & & & & & & & & & \\
\hspace{0.5cm}ElasticNet  & 0.47 & 0.44 & 0.46 & 0.34 & 0.31 & 5,592.71 & 0.21 & 0.28 & 12,148.00 \\
\hspace{0.5cm}Random Forest & 0.55 & 0.52 & 0.43 & 0.38 & 0.35 & 5,508.20 & 0.24 & 0.31 & 12,074.77 \\
\hspace{0.5cm}XGBoost  & 0.55 & 0.51 & 0.44 & 0.36 & 0.34 & 5,538.77 & 0.23 & 0.29 & 12,138.05 \\
Outdoor: & & & & & & & & &  \\
\hspace{0.5cm}ElasticNet & 0.69 & 0.69 & 0.39 & 0.46 & 0.43 & 5,288.28 & 0.33 & 0.34 & 11,778.49 \\
\hspace{0.5cm}Random Forest & 0.71 & 0.70 & 0.37 & 0.48 & 0.44 & 5,223.62 & 0.33 & 0.40 & 11,766.68 \\
\hspace{0.5cm}XGBoost & 0.70 & 0.69 & 0.38 & 0.47 & 0.43 & 5,243.45 & 0.33 & 0.38 & 11,804.78 \\
Complete: & & & & & & & & & \\
\hspace{0.5cm}ElasticNet & 0.84 & 0.84 & 0.31 & 0.59 & 0.53 & 4,794.78 & 0.42 & 0.48 & 11,374.23 \\
\hspace{0.5cm}Random Forest & 0.85 & 0.85 & 0.28 & 0.62 & 0.58 & 4,664.39 & 0.47 & 0.50 & 11,261.17 \\
\hspace{0.5cm}XGBoost & 0.84 & 0.84 & 0.28 & 0.61 & 0.57 & 4,765.31 & 0.39 & 0.48 & 11,536.06 \\
\hline
\end{tabular}
\end{table}

Prediction accuracy statistics for the same models but using feature vectors extracted from the original VGG16 network without fine-tuning were only slightly worse compared to the main results (supplementary Table S6). As an example, Spearman correlation coefficients in complete random forest models were 0.84, 0.55 and 0.45 for assets, expenditure, and income-based SEP when using the unmodified VGG16 CNN for feature extraction.

\subsection{Explaining and optimizing models}

\begin{figure}[ht!]
\centering 
\includegraphics[width=0.8\textwidth]{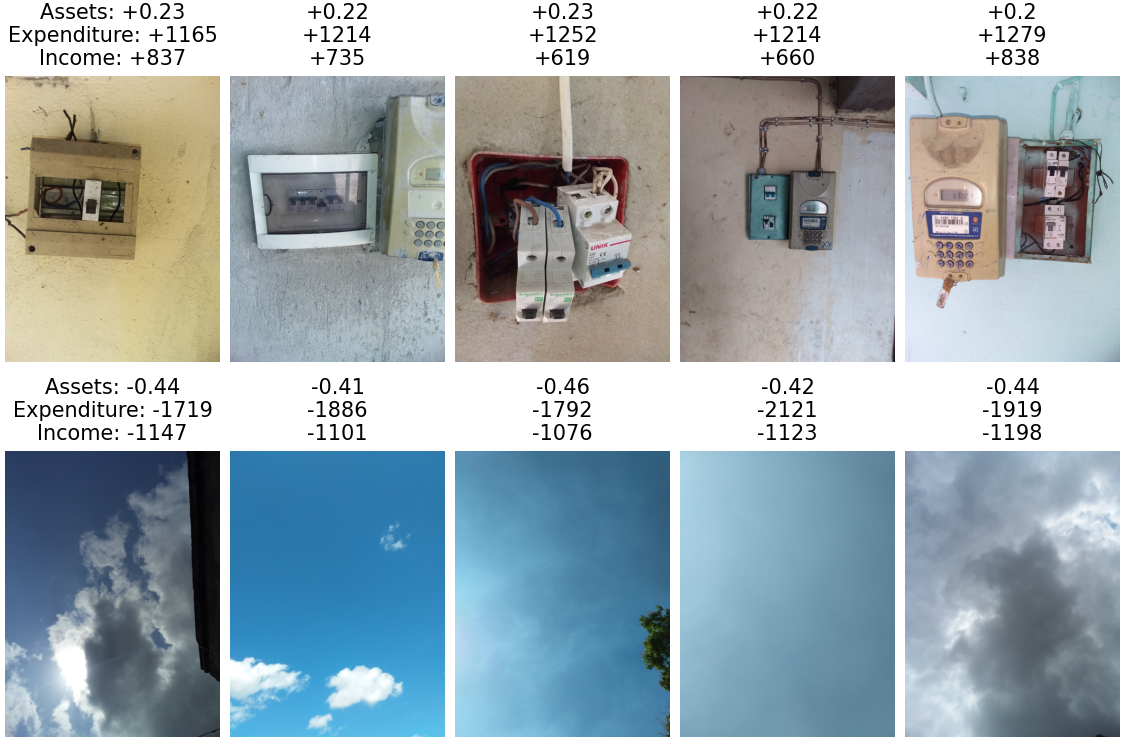}
\caption{Light source images with the largest positive (top row) and negative (bottom row) SHAP values in random forest complete models. Images were selected according to the top and bottom average SHAP value ranks across the three SEP measures.}
\label{fig:shap_lightsource}
\end{figure}

Inspection of the images with the largest positive and negative SHAP values in complete random forest models showed very clear patterns. In the case of the light source (Figure \ref{fig:shap_lightsource}), all images with positive SHAP values showed fuse boxes meaning that the household is connected to the electricity network, whereas negative SHAP values displayed the sky, signaling that the household only used natural light. Results for the rest of image types are available in supplementary Figures S8-S15: while different materials could be appreciated in floor images with the largest positive and negative SHAP values, different objects were present for water source, latrines, and stoves. Interestingly, for the roof category the object was the same (i.e., a corrugated roof) but differences in maintenance could be appreciated. The built environment was associated with positive SHAP values in street view and satellite images. Visualizations for the wall, front door, kitchen, and bathroom image types are not included due to potential risk of identification of the households.

\begin{figure}[!ht]
\centering 
\includegraphics[width=0.85\textwidth]{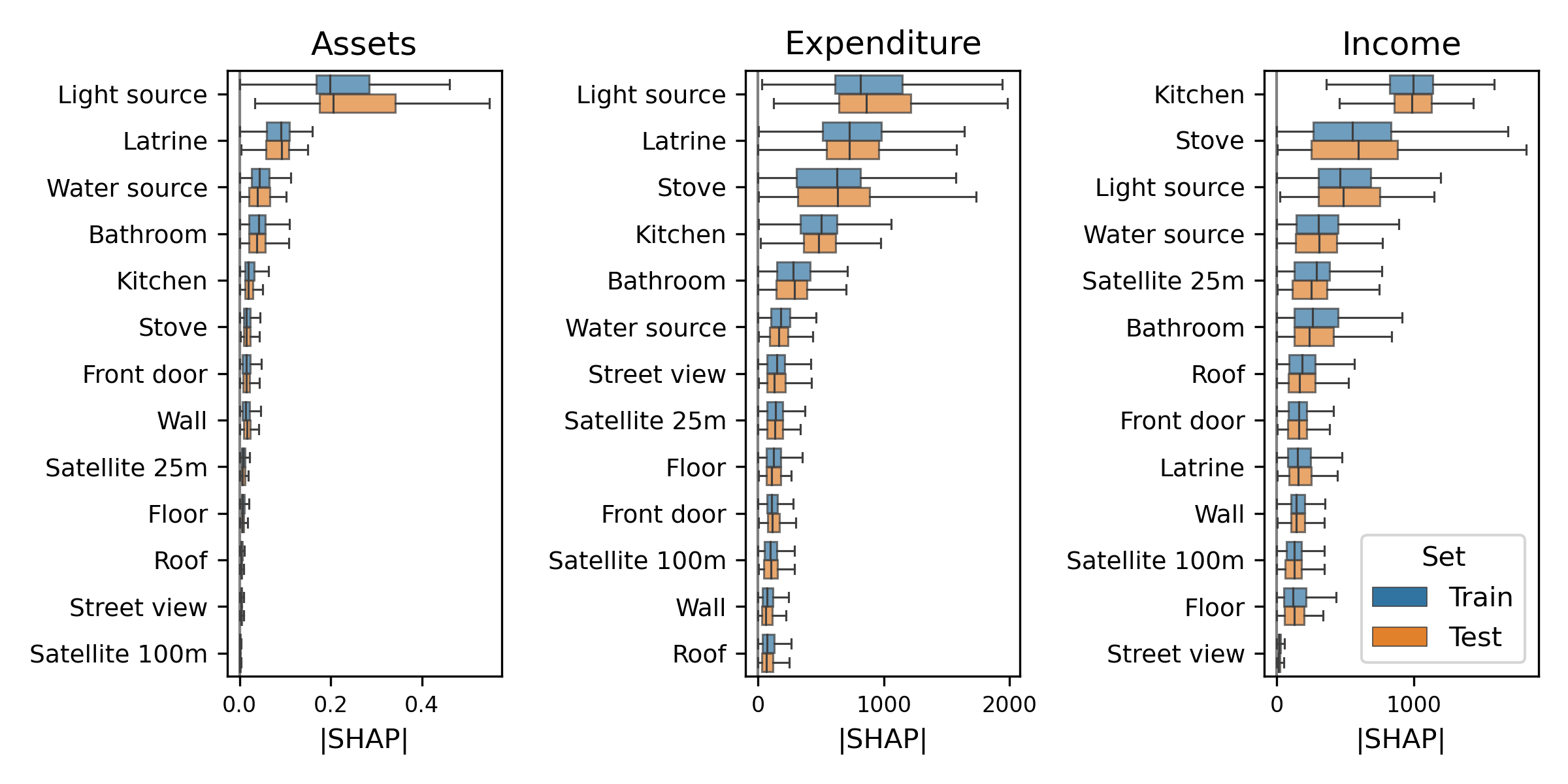}
\caption{Feature importance according to absolute SHAP values in complete random forest models, by SEP measure and train/test set. Image types are ordered according to the median absolute SHAP value.}
\label{fig:shap_importance}
\end{figure}

The most important image types in the best-performing random forest "complete" models were the light source for asset and expenditure-based SEP, and kitchen for income-based SEP (Figure \ref{fig:shap_importance}). Generally, indoor image types of specific items (e.g., light and water source, latrine) were more relevant than building materials (e.g., roof, floor, wall) and outdoor (front door, street view) images. Amongst satellite images, those with the smaller buffer size (25m) were more important.

The performance of "reduced" random forest regression models, which used the "outdoor" predictor set as well as the most important indoor image type identified in SHAP feature importance analyses, was higher than that of "outdoor" models (Table \ref{tab:XAIres}). Compared to "complete" models, the performance of "reduced" models was only slightly lower when modelling expenditure, whereas differences in assets and especially income-based SEP were somewhat larger (Table \ref{tab:XAIres}).

\begin{table}[!ht]
\caption{Prediction accuracy of outdoor, reduced, and complete random forest models measured with Pearson ($r$) and Spearman ($\rho$) correlation coefficients and RMSE in the test data by SEP measure. The added image type in reduced models is the light source for asset and expenditure-based SEP, and the kitchen for income-based SEP.}
\label{tab:XAIres}
\centering
\begin{tabular}{|l|ccc|ccc|ccc|}
\hline
Model &  & Assets & & & Expenditure & & & Income & \\
 & $r$ & $\rho$ & RMSE & $r$ & $\rho$ & RMSE & $r$ & $\rho$ & RMSE \\
\hline
Outdoor & 0.71 & 0.70 & 0.37 & 0.48 & 0.44 & 5,223.62 & 0.33 & 0.40 & 11,766.68 \\
Reduced & 0.79 & 0.80 & 0.32 & 0.62 & 0.51 & 4,754.23 & 0.36 & 0.39 & 11,748.99 \\
Complete & 0.85 & 0.85 & 0.28 & 0.62 & 0.58 & 4,664.39 & 0.47 & 0.50 & 11,261.17 \\
\hline
\end{tabular}
\end{table}

\newpage
\section{Discussion}

We assembled a dataset consisting of satellite imagery and a photograph survey of 11 outdoor and indoor household elements, which we used to predict asset, expenditure, and income-based household SEP in a semi-rural area of southern Mozambique. We found a moderate correlation between the three SEP measures. Feature maps were extracted from the images via transfer learning of a CNN. The best performance was found when modelling asset-based SEP in random forest models using features from all image types, while performance for expenditure and income-based SEP was lower. The performance of models only relying on satellite images was low, whereas the performance of a model comprising satellite, outdoor, and the most important household indoor image identified via SHAP was only slightly lower compared to models using all images. SHAP analyses also confirmed the reliability of the predictive models when inspecting the images with the largest effects, and helped gaining new insights.

The moderate correlations between the three SEP measures we found are aligned with the results of a synthesis study \citep{poirier2020}, which reported an average Spearman correlation across studies in LMICs of 0.55 between the SEP of assets and expenditure (we observed 0.63) and 0.42 between assets and income (we observed 0.47). These correlations can be partly explained by the different timing of each measure: while in rural areas of sub-Saharan Africa asset-based SEP is a mostly static measure that captures the slow wealth accumulation resulting from past saving behaviour over time, income has been found to have a low transmission to expenditure due to consumption insurance mechanisms \citep{2018Demagalhaes}. The higher correlation between asset and expenditure-based SEP may be explained by the ability of both measures to capture long-term economic status as opposed to income, which is more susceptible to short-term shocks \citep{Howe2009}. 

The modification and fine-tuning of a CNN did not translate into significantly better prediction accuracy compared to a model using off-the-shelf feature vectors extracted from a pre-trained model. We attribute this to the difference in tasks of the original network (i.e., ImageNet) vs. our application, which would require retraining a larger number of layers \citep{yosinski2014} that we could not perform due to sample size limitations. Past studies using satellite imagery to predict SEP have used satellite nighttime lights as a SEP proxy to increase sample size when training CNNs \citep{Jean2016, xie2016, head2017}. However, this is not a possibility in our study because the spatial resolution of these products is too coarse ($\sim$500 m for VIIRS-DNB \citep{Elvidge2017}) to characterize individual households. Another possibility would be to leverage a dataset closer to our task such as Dollar Street \citep{rojas2022}. However, we believe that its small sample size (the median sample size per image type in Dollar Street is 110, where only 20\% come from African countries \citep{rojas2022}) would not compensate for the heterogeneity between datasets. 

The prediction accuracy of our best-performing random forest model for asset-based SEP using all image types was comparable to the average performance found in the SEP mapping review by \citet{Hall2023} (the proportion of explained variance was 0.72 in our study vs. 0.75 in the review). That said, almost all studies included in the review predicted area-level SEP, which is likely less challenging than predicting SEP for individual households. The performance of expenditure and income models was substantially lower, also aligned with the SEP mapping literature \citep{Hall2023}. We attribute the better performance of asset models to the fact that our photograph survey focused on specific elements included in the asset-based SEP construction. The performance of satellite-only models was low, which is likely due to the limited variability of the built environment in the study area, with virtually all dwellings consisting of detached buildings. On the other hand, the increase in performance when adding ground-based photographs indicates that there are important factors that could not be captured with aerial imagery alone. Finally, outdoor models for asset-based SEP yielded relatively good results, indicating that studies predicting household asset-based measures could benefit from street view data in places where those data already exist \cite[e.g.,][]{suel2019}.

In SHAP analyzes, not only did we find clear differences in the items present in the photographs when comparing images with the largest positive and negative effects, but we also found differences in maintenance levels for the roof category. This finding supports the hypothesis that images have the potential to bypass the limitations of questionnaires when assessing asset quality \citep{Howe2012}. Furthermore, we discovered that a higher built environment is associated with higher SEP, which highlights the rural-(semi)urban differences in SEP in our population. Thanks to SHAP, we also found that only using one indoor household item was enough to achieve a performance close to that of models using all images. This finding opens the door to more efficient data collection, where outdoor photographs can be collected with minimal impact on the participant, and only one/few indoor household elements need to be included. Among indoor photographs, light source and kitchen were the two most important elements. We hypothesize this is because of the stark differences in these elements in our population, which might be different for other populations or in urban areas \citep{Howland2021}.

Our study has several strengths. First, we compiled a multimodal dataset that included satellite and ground-based outdoor and indoor household imagery which, to our knowledge, has never been done before. Second, we modelled and predicted individual household SEP, which is likely more challenging than estimating spatial aggregates and has been seldom attempted so far. Third, our analysis comprised three different SEP measures, reflecting the several ways SEP can be understood and measured. In that sense, the multi-output CNN we propose is an innovative way to extract feature vectors that refer to different yet related SEP measures. Finally, we used explainable machine learning as a means to balance performance and data collection burden in future studies, as well as to ensure that the models conform to prior knowledge.

Nonetheless, we also faced limitations. First, the low sample size was a limiting factor both during feature vector extraction (see CNN fine-tuning discussion) and supervised analyses, where the number of features was not much smaller than the number of samples. Second, our expenditure-based SEP measure did not contemplate non-monetary consumption prevalent in the study area, which we omitted due to the long consumption diaries needed to estimate it. However, we think that the impact of this is likely small given that simple expenditure questionnaires like the one we used have been found to correlate well with more complex monetary value methods \citep{Morris2000}. Lastly, we did not have data of a separate region to evaluate spatial model transferability.

Our dataset opens the door to additional research questions beyond SEP prediction. First, the collected household imagery captured many elements that affect health and therefore could be used to assess exposure and/or design interventions \citep{Weichental2019}: Stove and kitchen photographs for household air pollution; latrines, bathrooms, and water source for water quality and sanitation; wall, floor, and roof as housing quality indicators related to exposures such as indoor temperature. Second, image classification or object detection algorithms could be applied to automatically derive household asset information from the images (see for example \citet{Laranjeira2024} for building façade characteristics). This could lead to richer asset indicators compared to questionnaires, since images could also capture asset condition, quality, and surroundings \citep{Howe2012}. Lastly, methods such as Grad-CAM could be used to investigate the most important elements within the images \citep{abitbol2020}.

In this article, we showed how ground-based household photographs can complement satellite imagery for SEP prediction, allowing to zoom in from area-level predictions to individual households. The higher data collection effort compared to satellite-only approaches can be minimized using explainable machine learning. The developed prediction workflow can be potentially integrated into routine household surveys (e.g., DHS), where the collected imagery can also be used for other purposes such as refined asset characterization and environmental exposure assessment.

\section{Acknowledgments}

We would like to thank all the community members and local authorities who have supported and/or participated in the research. We also thank all the researchers and field staff involved in this project and all members of the board of patrons of the Manhiça Foundation, the legal representative of CISM for their support. 

The project "Novel methods to assess household socioeconomic position in Manhiça district in Mozambique" was funded through the Severo Ochoa Programme at ISGlobal. CISM is supported by the Government of Mozambique and the Spanish Agency for International Development (AECID). We acknowledge support from the grant CEX2023-0001290-S funded by MCIN/AEI/ 10.13039/501100011033, and support from the Generalitat de Catalunya through the CERCA Program.

\newpage
\bibliographystyle{unsrtnat}
\bibliography{references}  

\begin{thebibliography}{55}
\providecommand{\natexlab}[1]{#1}
\providecommand{\url}[1]{\texttt{#1}}
\expandafter\ifx\csname urlstyle\endcsname\relax
  \providecommand{\doi}[1]{doi: #1}\else
  \providecommand{\doi}{doi: \begingroup \urlstyle{rm}\Url}\fi

\bibitem[Galobardes et~al.(2006)Galobardes, Shaw, Lawlor, Lynch, and Davey~Smith]{Galobardes2006}
Bruna Galobardes, Mary Shaw, Debbie~A Lawlor, John~W Lynch, and George Davey~Smith.
\newblock Indicators of socioeconomic position (part 1).
\newblock \emph{Journal of Epidemiology \& Community Health}, 60\penalty0 (1):\penalty0 7--12, 2006.
\newblock ISSN 0143-005X.
\newblock \doi{10.1136/jech.2004.023531}.
\newblock URL \url{https://jech.bmj.com/content/60/1/7}.

\bibitem[Adler and Stewart(2010)]{Adler2010}
Nancy~E. Adler and Judith Stewart.
\newblock Health disparities across the lifespan: Meaning, methods, and mechanisms.
\newblock \emph{Annals of the New York Academy of Sciences}, 1186\penalty0 (1):\penalty0 5--23, 2010.
\newblock \doi{https://doi.org/10.1111/j.1749-6632.2009.05337.x}.
\newblock URL \url{https://nyaspubs.onlinelibrary.wiley.com/doi/abs/10.1111/j.1749-6632.2009.05337.x}.

\bibitem[Hystad et~al.(2019)Hystad, Duong, Brauer, Larkin, Arku, Kurmi, Fan, Avezum, Azam, Chifamba, Dans, du~Plessis, Gupta, Kumar, Lanas, Liu, Lu, Lopez-Jaramillo, Mony, Mohan, Mohan, Nair, Puoane, Rahman, Lap, Wang, Wei, Yeates, Rangarajan, Teo, Yusuf, and null null]{Hystad2019}
Perry Hystad, MyLinh Duong, Michael Brauer, Andrew Larkin, Raphael Arku, Om~P. Kurmi, Wen~Qi Fan, Alvaro Avezum, Igbal Azam, Jephat Chifamba, Antonio Dans, Johan~L. du~Plessis, Rajeev Gupta, Rajesh Kumar, Fernando Lanas, Zhiguang Liu, Yin Lu, Patricio Lopez-Jaramillo, Prem Mony, Viswanathan Mohan, Deepa Mohan, Sanjeev Nair, Thandi Puoane, Omar Rahman, Ah~Tse Lap, Yanga Wang, Li~Wei, Karen Yeates, Sumathy Rangarajan, Koon Teo, Salim Yusuf, and null null.
\newblock Health effects of household solid fuel use: Findings from 11 countries within the prospective urban and rural epidemiology study.
\newblock \emph{Environmental Health Perspectives}, 127\penalty0 (5):\penalty0 057003, 2019.
\newblock \doi{10.1289/EHP3915}.
\newblock URL \url{https://ehp.niehs.nih.gov/doi/abs/10.1289/EHP3915}.

\bibitem[Filmer and Pritchett(2001)]{Filmer2001}
Deon Filmer and Lant~H. Pritchett.
\newblock {Estimating wealth effects without expenditure data—or tears: An application to educational enrollments in states of India}.
\newblock \emph{Demography}, 38\penalty0 (1):\penalty0 115--132, 02 2001.
\newblock \doi{10.1353/dem.2001.0003}.
\newblock URL \url{https://doi.org/10.1353/dem.2001.0003}.

\bibitem[Filmer and Scott(2012)]{filmer2012}
Deon Filmer and Kinnon Scott.
\newblock Assessing asset indices.
\newblock \emph{Demography}, 49:\penalty0 359--392, 2012.

\bibitem[Howe et~al.(2012)Howe, Galobardes, Matijasevich, Gordon, Johnston, Onwujekwe, Patel, Webb, Lawlor, and Hargreaves]{Howe2012}
Laura~D Howe, Bruna Galobardes, Alicia Matijasevich, David Gordon, Deborah Johnston, Obinna Onwujekwe, Rita Patel, Elizabeth~A Webb, Debbie~A Lawlor, and James~R Hargreaves.
\newblock {Measuring socio-economic position for epidemiological studies in low- and middle-income countries: a methods of measurement in epidemiology paper}.
\newblock \emph{International Journal of Epidemiology}, 41\penalty0 (3):\penalty0 871--886, 03 2012.
\newblock ISSN 0300-5771.
\newblock \doi{10.1093/ije/dys037}.
\newblock URL \url{https://doi.org/10.1093/ije/dys037}.

\bibitem[Olivia~Howland and Brockington(2021)]{Howland2021}
Christine~Noe Olivia~Howland and Dan Brockington.
\newblock The multiple meanings of prosperity and poverty: a cross-site comparison from tanzania.
\newblock \emph{The Journal of Peasant Studies}, 48\penalty0 (1):\penalty0 180--200, 2021.
\newblock \doi{10.1080/03066150.2019.1658080}.
\newblock URL \url{https://doi.org/10.1080/03066150.2019.1658080}.

\bibitem[Bai et~al.(2020)Bai, Naumova, and Masters]{Bai2020}
Yan Bai, Elena~N. Naumova, and William~A. Masters.
\newblock Seasonality of diet costs reveals food system performance in east africa.
\newblock \emph{Science Advances}, 6\penalty0 (49):\penalty0 eabc2162, 2020.
\newblock \doi{10.1126/sciadv.abc2162}.
\newblock URL \url{https://www.science.org/doi/abs/10.1126/sciadv.abc2162}.

\bibitem[Burke et~al.(2021)Burke, Driscoll, Lobell, and Ermon]{Burke2021}
Marshall Burke, Anne Driscoll, David~B. Lobell, and Stefano Ermon.
\newblock Using satellite imagery to understand and promote sustainable development.
\newblock \emph{Science}, 371\penalty0 (6535):\penalty0 eabe8628, 2021.
\newblock \doi{10.1126/science.abe8628}.
\newblock URL \url{https://www.science.org/doi/abs/10.1126/science.abe8628}.

\bibitem[Gao et~al.(2019)Gao, Zhang, and Zhou]{Gao2019}
Jian Gao, Yi-Cheng Zhang, and Tao Zhou.
\newblock Computational socioeconomics.
\newblock \emph{Physics Reports}, 817:\penalty0 1--104, 2019.
\newblock ISSN 0370-1573.
\newblock \doi{https://doi.org/10.1016/j.physrep.2019.05.002}.
\newblock URL \url{https://www.sciencedirect.com/science/article/pii/S0370157319301954}.
\newblock Computational Socioeconomics.

\bibitem[Hall et~al.(2023)Hall, Dompae, Wahab, and Dzanku]{Hall2023}
Ola Hall, Francis Dompae, Ibrahim Wahab, and Fred~Mawunyo Dzanku.
\newblock A review of machine learning and satellite imagery for poverty prediction: Implications for development research and applications.
\newblock \emph{Journal of International Development}, 35\penalty0 (7):\penalty0 1753--1768, 2023.
\newblock \doi{https://doi.org/10.1002/jid.3751}.
\newblock URL \url{https://onlinelibrary.wiley.com/doi/abs/10.1002/jid.3751}.

\bibitem[Elvidge et~al.(2009)Elvidge, Sutton, Ghosh, Tuttle, Baugh, Bhaduri, and Bright]{Elvidge2009}
Christopher~D. Elvidge, Paul~C. Sutton, Tilottama Ghosh, Benjamin~T. Tuttle, Kimberly~E. Baugh, Budhendra Bhaduri, and Edward Bright.
\newblock A global poverty map derived from satellite data.
\newblock \emph{Computers \& Geosciences}, 35\penalty0 (8):\penalty0 1652--1660, 2009.
\newblock ISSN 0098-3004.
\newblock \doi{https://doi.org/10.1016/j.cageo.2009.01.009}.
\newblock URL \url{https://www.sciencedirect.com/science/article/pii/S0098300409001253}.

\bibitem[Head et~al.(2017)Head, Manguin, Tran, and Blumenstock]{head2017}
Andrew Head, M{\'e}lanie Manguin, Nhat Tran, and Joshua~E Blumenstock.
\newblock Can human development be measured with satellite imagery?
\newblock \emph{Ictd}, 17:\penalty0 16--19, 2017.

\bibitem[Sohnesen et~al.(2022)Sohnesen, Fisker, and Malmgren-Hansen]{Sohnesen2022}
Thomas~Pave Sohnesen, Peter Fisker, and David Malmgren-Hansen.
\newblock Using satellite data to guide urban poverty reduction.
\newblock \emph{Review of Income and Wealth}, 68\penalty0 (S2):\penalty0 S282--S294, 2022.
\newblock \doi{https://doi.org/10.1111/roiw.12552}.
\newblock URL \url{https://onlinelibrary.wiley.com/doi/abs/10.1111/roiw.12552}.

\bibitem[Yeh et~al.(2020)Yeh, Perez, Driscoll, Azzari, Tang, Lobell, Ermon, and Burke]{Yeh2020}
Christopher Yeh, Anthony Perez, Anne Driscoll, George Azzari, Zhongyi Tang, David Lobell, Stefano Ermon, and Marshall Burke.
\newblock Using publicly available satellite imagery and deep learning to understand economic well-being in africa.
\newblock \emph{Nature communications}, 11\penalty0 (1):\penalty0 2583, 2020.

\bibitem[Jonathan~Hersh and Mann(2021)]{Hersh2021}
Ryan~Engstrom Jonathan~Hersh and Michael Mann.
\newblock Open data for algorithms: mapping poverty in belize using open satellite derived features and machine learning.
\newblock \emph{Information Technology for Development}, 27\penalty0 (2):\penalty0 263--292, 2021.
\newblock \doi{10.1080/02681102.2020.1811945}.
\newblock URL \url{https://doi.org/10.1080/02681102.2020.1811945}.

\bibitem[Jean et~al.(2016)Jean, Burke, Xie, Davis, Lobell, and Ermon]{Jean2016}
Neal Jean, Marshall Burke, Michael Xie, W.~Matthew Davis, David~B. Lobell, and Stefano Ermon.
\newblock Combining satellite imagery and machine learning to predict poverty.
\newblock \emph{Science}, 353\penalty0 (6301):\penalty0 790--794, 2016.
\newblock \doi{10.1126/science.aaf7894}.
\newblock URL \url{https://www.science.org/doi/abs/10.1126/science.aaf7894}.

\bibitem[Zhao et~al.(2019)Zhao, Yu, Liu, Chen, Li, Wang, and Wu]{zhao2020}
Xizhi Zhao, Bailang Yu, Yan Liu, Zuoqi Chen, Qiaoxuan Li, Congxiao Wang, and Jianping Wu.
\newblock Estimation of poverty using random forest regression with multi-source data: A case study in bangladesh.
\newblock \emph{Remote Sensing}, 11\penalty0 (4), 2019.
\newblock ISSN 2072-4292.
\newblock \doi{10.3390/rs11040375}.
\newblock URL \url{https://www.mdpi.com/2072-4292/11/4/375}.

\bibitem[Ayush et~al.(2020)Ayush, Uzkent, Burke, Lobell, and Ermon]{Ayush2020}
Kumar Ayush, Burak Uzkent, Marshall Burke, David Lobell, and Stefano Ermon.
\newblock Generating interpretable poverty maps using object detection in satellite images.
\newblock \emph{arXiv preprint}, 2020.
\newblock URL \url{https://arxiv.org/abs/2002.01612}.

\bibitem[Engstrom et~al.(2022)Engstrom, Hersh, and Newhouse]{engstrom2022}
Ryan Engstrom, Jonathan Hersh, and David Newhouse.
\newblock Poverty from space: Using high resolution satellite imagery for estimating economic well-being.
\newblock \emph{The World Bank Economic Review}, 36\penalty0 (2):\penalty0 382--412, 2022.

\bibitem[Niu et~al.(2020)Niu, Chen, and Yuan]{Niu2020}
Tong Niu, Yimin Chen, and Yuan Yuan.
\newblock Measuring urban poverty using multi-source data and a random forest algorithm: A case study in guangzhou.
\newblock \emph{Sustainable Cities and Society}, 54:\penalty0 102014, 2020.
\newblock ISSN 2210-6707.
\newblock \doi{https://doi.org/10.1016/j.scs.2020.102014}.
\newblock URL \url{https://www.sciencedirect.com/science/article/pii/S2210670720300019}.

\bibitem[Duque et~al.(2015)Duque, Patino, Ruiz, and Pardo-Pascual]{Duque2015}
Juan~C. Duque, Jorge~E. Patino, Luis~A. Ruiz, and Josep~E. Pardo-Pascual.
\newblock Measuring intra-urban poverty using land cover and texture metrics derived from remote sensing data.
\newblock \emph{Landscape and Urban Planning}, 135:\penalty0 11--21, 2015.
\newblock ISSN 0169-2046.
\newblock \doi{https://doi.org/10.1016/j.landurbplan.2014.11.009}.
\newblock URL \url{https://www.sciencedirect.com/science/article/pii/S0169204614002692}.

\bibitem[Xie et~al.(2016)Xie, Jean, Burke, Lobell, and Ermon]{xie2016}
Michael Xie, Neal Jean, Marshall Burke, David Lobell, and Stefano Ermon.
\newblock Transfer learning from deep features for remote sensing and poverty mapping.
\newblock In \emph{Proceedings of the AAAI conference on artificial intelligence}, volume~30, 2016.

\bibitem[Chi et~al.(2022)Chi, Fang, Chatterjee, and Blumenstock]{chi2022}
Guanghua Chi, Han Fang, Sourav Chatterjee, and Joshua~E. Blumenstock.
\newblock Microestimates of wealth for all low- and middle-income countries.
\newblock \emph{Proceedings of the National Academy of Sciences}, 119\penalty0 (3):\penalty0 e2113658119, 2022.
\newblock \doi{10.1073/pnas.2113658119}.
\newblock URL \url{https://www.pnas.org/doi/abs/10.1073/pnas.2113658119}.

\bibitem[Watmough et~al.(2019)Watmough, Marcinko, Sullivan, Tschirhart, Mutuo, Palm, and Svenning]{watmough2019}
Gary~R Watmough, Charlotte~LJ Marcinko, Clare Sullivan, Kevin Tschirhart, Patrick~K Mutuo, Cheryl~A Palm, and Jens-Christian Svenning.
\newblock Socioecologically informed use of remote sensing data to predict rural household poverty.
\newblock \emph{Proceedings of the National Academy of Sciences}, 116\penalty0 (4):\penalty0 1213--1218, 2019.

\bibitem[McBride et~al.(2022)McBride, Barrett, Browne, Hu, Liu, Matteson, Sun, and Wen]{mcbride2022}
Linden McBride, Christopher~B. Barrett, Christopher Browne, Leiqiu Hu, Yanyan Liu, David~S. Matteson, Ying Sun, and Jiaming Wen.
\newblock Predicting poverty and malnutrition for targeting, mapping, monitoring, and early warning.
\newblock \emph{Applied Economic Perspectives and Policy}, 44\penalty0 (2):\penalty0 879--892, 2022.
\newblock \doi{https://doi.org/10.1002/aepp.13175}.
\newblock URL \url{https://onlinelibrary.wiley.com/doi/abs/10.1002/aepp.13175}.

\bibitem[Gebru et~al.(2017)Gebru, Krause, Wang, Chen, Deng, Aiden, and Fei-Fei]{Gebru2017}
Timnit Gebru, Jonathan Krause, Yilun Wang, Duyun Chen, Jia Deng, Erez~Lieberman Aiden, and Li~Fei-Fei.
\newblock Using deep learning and google street view to estimate the demographic makeup of neighborhoods across the united states.
\newblock \emph{Proceedings of the National Academy of Sciences}, 114\penalty0 (50):\penalty0 13108--13113, 2017.
\newblock \doi{10.1073/pnas.1700035114}.
\newblock URL \url{https://www.pnas.org/doi/abs/10.1073/pnas.1700035114}.

\bibitem[Suel et~al.(2019)Suel, Polak, Bennett, and Ezzati]{suel2019}
Esra Suel, John~W Polak, James~E Bennett, and Majid Ezzati.
\newblock Measuring social, environmental and health inequalities using deep learning and street imagery.
\newblock \emph{Scientific reports}, 9\penalty0 (1):\penalty0 6229, 2019.

\bibitem[Fan et~al.(2023)Fan, Zhang, Loo, and Ratti]{Fan2023}
Zhuangyuan Fan, Fan Zhang, Becky P.~Y. Loo, and Carlo Ratti.
\newblock Urban visual intelligence: Uncovering hidden city profiles with street view images.
\newblock \emph{Proceedings of the National Academy of Sciences}, 120\penalty0 (27):\penalty0 e2220417120, 2023.
\newblock \doi{10.1073/pnas.2220417120}.
\newblock URL \url{https://www.pnas.org/doi/abs/10.1073/pnas.2220417120}.

\bibitem[Rojas et~al.(2022)Rojas, Diamos, Kini, Kanter, Reddi, and Coleman]{rojas2022}
William A~Gaviria Rojas, Sudnya Diamos, Keertan~Ranjan Kini, David Kanter, Vijay~Janapa Reddi, and Cody Coleman.
\newblock The dollar street dataset: Images representing the geographic and socioeconomic diversity of the world.
\newblock In \emph{Thirty-sixth Conference on Neural Information Processing Systems Datasets and Benchmarks Track}, 2022.

\bibitem[Hall et~al.(2022)Hall, Ohlsson, and R{\"o}gnvaldsson]{hall2022}
Ola Hall, Mattias Ohlsson, and Thorsteinn R{\"o}gnvaldsson.
\newblock A review of explainable ai in the satellite data, deep machine learning, and human poverty domain.
\newblock \emph{Patterns}, 3\penalty0 (10), 2022.

\bibitem[Abitbol and Karsai(2020)]{abitbol2020}
Jacob~Levy Abitbol and Marton Karsai.
\newblock Interpretable socioeconomic status inference from aerial imagery through urban patterns.
\newblock \emph{Nature Machine Intelligence}, 2\penalty0 (11):\penalty0 684--692, 2020.

\bibitem[Nhacolo et~al.(2021)Nhacolo, Jamisse, Augusto, Matsena, Hunguana, Mandomando, Arnaldo, Munguambe, Macete, Alonso, Saúte, and Sacoor]{Nhacolo2021}
Ariel Nhacolo, Edgar Jamisse, Orvalho Augusto, Teodomiro Matsena, Aura Hunguana, Inácio Mandomando, Carlos Arnaldo, Khátia Munguambe, Eusébio Macete, Pedro Alonso, Francisco Saúte, and Charfudin Sacoor.
\newblock {Cohort Profile Update: Manhiça Health and Demographic Surveillance System (HDSS) of the Manhiça Health Research Centre (CISM)}.
\newblock \emph{International Journal of Epidemiology}, 50\penalty0 (2):\penalty0 395--395, 01 2021.
\newblock ISSN 0300-5771.
\newblock \doi{10.1093/ije/dyaa218}.
\newblock URL \url{https://doi.org/10.1093/ije/dyaa218}.

\bibitem[Oseni et~al.(2021)Oseni, Palacios-Lopez, Mugera, and Durazo]{oseni2021}
Gbemisola Oseni, Amparo Palacios-Lopez, Harriet~Kasidi Mugera, and Josefine Durazo.
\newblock \emph{Capturing What Matters: Essential Guidelines for Designing Household Surveys}.
\newblock World Bank, 2021.

\bibitem[Grau-Pujol et~al.(2022)Grau-Pujol, Cano, Marti-Soler, Casellas, Giorgi, Nhacolo, Saute, Gin{\'e}, Quint{\'o}, Sacoor, et~al.]{grau2022}
Berta Grau-Pujol, Jorge Cano, Helena Marti-Soler, Aina Casellas, Emanuele Giorgi, Ariel Nhacolo, Francisco Saute, Ricard Gin{\'e}, Lloren{\c{c}} Quint{\'o}, Charfudin Sacoor, et~al.
\newblock Neighbors’ use of water and sanitation facilities can affect children’s health: a cohort study in mozambique using a spatial approach.
\newblock \emph{BMC Public Health}, 22\penalty0 (1):\penalty0 983, 2022.

\bibitem[Poirier(2024)]{poirier2024}
Mathieu~JP Poirier.
\newblock Systematic comparison of household income, consumption, and assets to measure health inequalities in low-and middle-income countries.
\newblock \emph{Scientific Reports}, 14\penalty0 (1):\penalty0 3851, 2024.

\bibitem[Traissac and Martin-Prevel(2012)]{Traissac2012}
Pierre Traissac and Yves Martin-Prevel.
\newblock {Alternatives to principal components analysis to derive asset-based indices to measure socio-economic position in low- and middle-income countries: the case for multiple correspondence analysis}.
\newblock \emph{International Journal of Epidemiology}, 41\penalty0 (4):\penalty0 1207--1208, 08 2012.
\newblock ISSN 0300-5771.
\newblock \doi{10.1093/ije/dys122}.
\newblock URL \url{https://doi.org/10.1093/ije/dys122}.

\bibitem[Elgendy(2020)]{elgendy2020}
Mohamed Elgendy.
\newblock \emph{Deep learning for vision systems}.
\newblock Simon and Schuster, 2020.

\bibitem[Simonyan and Zisserman(2014)]{simonyan2014}
Karen Simonyan and Andrew Zisserman.
\newblock Very deep convolutional networks for large-scale image recognition.
\newblock \emph{arXiv preprint arXiv:1409.1556}, 2014.

\bibitem[Sharif~Razavian et~al.(2014)Sharif~Razavian, Azizpour, Sullivan, and Carlsson]{Razavian2014}
Ali Sharif~Razavian, Hossein Azizpour, Josephine Sullivan, and Stefan Carlsson.
\newblock Cnn features off-the-shelf: An astounding baseline for recognition.
\newblock In \emph{Proceedings of the IEEE Conference on Computer Vision and Pattern Recognition (CVPR) Workshops}, June 2014.

\bibitem[Zou and Hastie(2005)]{hou2005}
Hui Zou and Trevor Hastie.
\newblock {Regularization and Variable Selection Via the Elastic Net}.
\newblock \emph{Journal of the Royal Statistical Society Series B: Statistical Methodology}, 67\penalty0 (2):\penalty0 301--320, 03 2005.
\newblock ISSN 1369-7412.
\newblock \doi{10.1111/j.1467-9868.2005.00503.x}.
\newblock URL \url{https://doi.org/10.1111/j.1467-9868.2005.00503.x}.

\bibitem[Breiman(2001)]{breiman2001}
Leo Breiman.
\newblock Random forests.
\newblock \emph{Machine learning}, 45:\penalty0 5--32, 2001.

\bibitem[Chen and Guestrin(2016)]{xgboostmod}
Tianqi Chen and Carlos Guestrin.
\newblock Xgboost: A scalable tree boosting system.
\newblock In \emph{Proceedings of the 22nd ACM SIGKDD International Conference on Knowledge Discovery and Data Mining}, KDD '16, page 785–794, New York, NY, USA, 2016. Association for Computing Machinery.
\newblock ISBN 9781450342322.
\newblock \doi{10.1145/2939672.2939785}.
\newblock URL \url{https://doi.org/10.1145/2939672.2939785}.

\bibitem[Molnar(2020)]{molnar2020}
Christoph Molnar.
\newblock \emph{Interpretable machine learning}.
\newblock Lulu. com, 2020.

\bibitem[Lundberg et~al.(2020)Lundberg, Erion, Chen, DeGrave, Prutkin, Nair, Katz, Himmelfarb, Bansal, and Lee]{lundberg2020}
Scott~M. Lundberg, Gabriel Erion, Hugh Chen, Alex DeGrave, Jordan~M. Prutkin, Bala Nair, Ronit Katz, Jonathan Himmelfarb, Nisha Bansal, and Su-In Lee.
\newblock From local explanations to global understanding with explainable ai for trees.
\newblock \emph{Nature Machine Intelligence}, 2\penalty0 (1):\penalty0 2522--5839, 2020.

\bibitem[{R Core Team}(2022)]{R}
{R Core Team}.
\newblock \emph{R: A Language and Environment for Statistical Computing}.
\newblock R Foundation for Statistical Computing, Vienna, Austria, 2022.
\newblock URL \url{https://www.R-project.org/}.

\bibitem[Van~Rossum and Drake(2009)]{python3}
Guido Van~Rossum and Fred~L. Drake.
\newblock \emph{Python 3 Reference Manual}.
\newblock CreateSpace, Scotts Valley, CA, 2009.
\newblock ISBN 1441412697.

\bibitem[Poirier et~al.(2020)Poirier, Gr{\'e}pin, and Grignon]{poirier2020}
Mathieu~JP Poirier, Karen~A Gr{\'e}pin, and Michel Grignon.
\newblock Approaches and alternatives to the wealth index to measure socioeconomic status using survey data: a critical interpretive synthesis.
\newblock \emph{Social Indicators Research}, 148\penalty0 (1):\penalty0 1--46, 2020.

\bibitem[{De Magalhães} and Santaeulàlia-Llopis(2018)]{2018Demagalhaes}
Leandro {De Magalhães} and Raül Santaeulàlia-Llopis.
\newblock The consumption, income, and wealth of the poorest: An empirical analysis of economic inequality in rural and urban sub-saharan africa for macroeconomists.
\newblock \emph{Journal of Development Economics}, 134:\penalty0 350--371, 2018.
\newblock ISSN 0304-3878.
\newblock \doi{https://doi.org/10.1016/j.jdeveco.2018.05.014}.
\newblock URL \url{https://www.sciencedirect.com/science/article/pii/S0304387818305017}.

\bibitem[Howe et~al.(2009)Howe, Hargreaves, Gabrysch, and Huttly]{Howe2009}
L~D Howe, J~R Hargreaves, S~Gabrysch, and S~R~A Huttly.
\newblock Is the wealth index a proxy for consumption expenditure? a systematic review.
\newblock \emph{Journal of Epidemiology \& Community Health}, 63\penalty0 (11):\penalty0 871--877, 2009.
\newblock ISSN 0143-005X.
\newblock \doi{10.1136/jech.2009.088021}.
\newblock URL \url{https://jech.bmj.com/content/63/11/871}.

\bibitem[Yosinski et~al.(2014)Yosinski, Clune, Bengio, and Lipson]{yosinski2014}
Jason Yosinski, Jeff Clune, Yoshua Bengio, and Hod Lipson.
\newblock How transferable are features in deep neural networks?
\newblock \emph{Advances in neural information processing systems}, 27, 2014.

\bibitem[Elvidge et~al.(2017)Elvidge, Baugh, Zhizhin, Hsu, and Ghosh]{Elvidge2017}
Christopher~D Elvidge, Kimberly Baugh, Mikhail Zhizhin, Feng~Chi Hsu, and Tilottama Ghosh.
\newblock Viirs night-time lights.
\newblock \emph{International journal of remote sensing}, 38\penalty0 (21):\penalty0 5860--5879, 2017.

\bibitem[Morris et~al.(2000)Morris, Carletto, Hoddinott, and Christiaensen]{Morris2000}
Saul~S Morris, Calogero Carletto, John Hoddinott, and Luc J~M Christiaensen.
\newblock Validity of rapid estimates of household wealth and income for health surveys in rural africa.
\newblock \emph{Journal of Epidemiology \& Community Health}, 54\penalty0 (5):\penalty0 381--387, 2000.
\newblock ISSN 0143-005X.
\newblock \doi{10.1136/jech.54.5.381}.
\newblock URL \url{https://jech.bmj.com/content/54/5/381}.

\bibitem[Weichenthal et~al.(2019)Weichenthal, Hatzopoulou, and Brauer]{Weichental2019}
Scott Weichenthal, Marianne Hatzopoulou, and Michael Brauer.
\newblock A picture tells a thousand…exposures: Opportunities and challenges of deep learning image analyses in exposure science and environmental epidemiology.
\newblock \emph{Environment International}, 122:\penalty0 3--10, 2019.
\newblock ISSN 0160-4120.
\newblock \doi{https://doi.org/10.1016/j.envint.2018.11.042}.
\newblock URL \url{https://www.sciencedirect.com/science/article/pii/S0160412018322001}.

\bibitem[Laranjeira et~al.(2024)Laranjeira, Pereira, Oliveira, Barbosa, Fernandes, Bermudi, Resende, Fernandes, Nogueira, Andrade, Quintanilha, dos Santos, and Chiaravalloti-Neto]{Laranjeira2024}
Camila Laranjeira, Matheus Pereira, Raul Oliveira, Gerson Barbosa, Camila Fernandes, Patricia Bermudi, Ester Resende, Eduardo Fernandes, Keiller Nogueira, Valmir Andrade, José~Alberto Quintanilha, Jefersson~A. dos Santos, and Francisco Chiaravalloti-Neto.
\newblock Automatic mapping of high-risk urban areas for aedes aegypti infestation based on building facade image analysis.
\newblock \emph{PLOS Neglected Tropical Diseases}, 18\penalty0 (6):\penalty0 1--29, 06 2024.
\newblock \doi{10.1371/journal.pntd.0011811}.
\newblock URL \url{https://doi.org/10.1371/journal.pntd.0011811}.

\end{thebibliography}

\end{document}


\maketitle

\newpage
\textbf{Supplementary Methods S1:} Hyperparameter search of the CNN.

We performed a set of experiments to determine the most suitable values for the model hyperparameters by defining a set of candidate values for each hyperparameter and then performing a random search with 10 iterations. To do so, we randomly divided the 800 households in the training data into two groups. The first had 640 households and was used to train the models (50 epochs), and the second had 160 households and was used to evaluate the results of the trained models with the different hyperparameter combinations (see the table below for hyperparameter candidates). The remaining 175 households in the test data set were not used at any point to avoid leakage. During training, we monitored the learning progress by computing, at the end of each epoch, the average loss per batch and the percentage accuracy for each of the binary SEP measures. We performed this analysis for four image types out of the total thirteen: floor, stove, front door, and satellite (25m). We chose the parameter combination that yielded the lowest error in the evaluation group while having a stable loss and minimal signs of overfitting in the four image types we tested (see the table below for the final parameter values). We assumed that these parameter values were also appropriate for all image types.

\begin{table}[h]
    \centering
    \begin{tabular}{c|c|c}
        \hline
        Parameter & Candidate values & Chosen value \\
        \hline
        Learning rate & 1e-2, 1e-3, 1e-4 & 1e-2 \\
        Batch size & 8, 16, 32, 64 & 32 \\
        L2 regularization & 1e-1, 1e-2, 1e-4, 1e-6 & 1e-4 \\
        Momentum & 0.5, 0.9, 0.99 & 0.5 \\
        \hline
    \end{tabular}
\end{table}

\newpage
\begin{table}[!h]
\caption{List of income and expenditure sources included in the questionnaires and used in the construction of the SEP measures. All quantities were asked in terms of monthly mean income/expenditure in metical.}
\centering
\begin{tabular}{ll}
\hline
\textbf{Measure} & \textbf{Source}\\
\hline
Income & Self-employed (owned business or agricultural activity)  \\ 
Income & Salaries, i.e. employed by a person outside the household \\ 
Income & Other income sources (renting or selling actives, subsidies) \\ 
Expenditure & Food and beverages \\  
Expenditure & Cooking and lighting fuel \\  
Expenditure & Health (insurance, medication, treatments) \\  
Expenditure & Electricity \\  
Expenditure & Education (registration and equipment) \\  
Expenditure & Agricultural equipment (tools, seeds, pesticides, fertilizers) \\  
Expenditure & Clothing \\  
Expenditure & Cleaning and hygiene products \\  
Expenditure & Household tools and appliances (e.g. cooking ware, small devices, water containers) \\ 
Expenditure & Insurance (health, life, vehicle) \\
Expenditure & Transport (public transport fares, fuel, reparations, bicycles) \\
Expenditure & Hobbies (e.g. sport equipment, books) \\
\hline
\end{tabular}
\end{table}

\newpage
\begin{table}[!h]
\caption{List of assets included in the questionnaires and used in the construction of the asset-based SEP measure. Note that some categories were collapsed due to sample size limitations.}
\centering
\begin{tabular}{ll}
\hline
\textbf{Variable}	& \textbf{Categories} \\
\hline
Owns a television & 1) Yes, 2) no \\ 
Owns a radio & 1) Yes, 2) no \\ 
Owns a computer & 1) Yes, 2) no \\ 
Owns a glacier, fridge or freezer & 1) Yes, 2) no \\ 
Owns a mobile or fixed telephone & 1) Yes, 2) no \\ 
Owns a motorcycle or car & 1) Yes, 2) no \\ 
Owns a bicycle & 1) Yes, 2) no \\ 
Owns a commercial farm & 1) Yes, 2) no \\ 
Owns pigs, goats or cattle & 1) Yes, 2) no \\ 
Type of wall & 1) Concrete, bricks, 2) wood, zinc, adobe, reeds, palms, others \\ 
Type of floor & 1) Concrete, 2) wood, marble, tiles, 3) adobe, dirt, others \\ 
Type of lighting fuel & 1) Electricity, generator, solar panel, 2) gas, petrol, candles, \\
& batteries, firewood, others  \\ 
Type of kitchen & 1) Indoors, 2) outdoors covered, 3) outdoors uncovered, \\ 
& 4) no kitchen \\ 
Type of kitchen fuel & 1) Gas, electricity, petrol, 2) charcoal, 3) wood, none, others \\ 
Type of water source & 1) Piped in the house, 2) piped in compound 3) non-piped  \\ 
Type of latrine & 1) Improved, 2) unimproved, 3) no latrine \\ 
\hline
\end{tabular}
\end{table}

\newpage
\begin{table}[!h]
\caption{Number of missing entries per image type.}
\centering
\begin{tabular}{lcc}
\hline
Image type & Count & \% of total samples \\
\hline
Roof & 11 & 1.1 \\
Floor & 3 & 0.3 \\
Light source & 16 & 1.6 \\
Front door & 5 & 0.5 \\
Wall & 3 & 0.3 \\
Kitchen & 39 & 4.0 \\
Stove & 7 & 0.7 \\
Bathroom & 7 & 0.7 \\
Latrine & 9 & 0.9 \\
Street view & 1 & 0.1 \\
Water source & 3 & 0.3 \\
Satellite (25m buffer) & 0 & 0.0 \\
Satellite (100m buffer) & 0 & 0.0 \\
\hline
\end{tabular}
\end{table}

\newpage
\begin{table}[!h]
\caption{List of hyperparameters tuned in the regression pipelines and their candidate values. The functions and hyperparameter names are listed according to \texttt{scikit-learn} and \texttt{xgboost} python libraries.}
\centering
\begin{tabular}{lll}
\hline
Function & Parameter & Candidate values \\
\hline
ElasticNet & alpha & 1, 1.5, 2 \\
& l1\_ratio & 0.1, 0.25, 0.4, 0.5, 0.6, 0.75, 0.9 \\
RandomForestRegressor$^1$ & min\_samples\_leaf & 20, 30, 40, 50 \\
 & max\_features & 0.1, 0.2, 0.3, 0.4, 0.5, 0.6, 0.7 \\
XGBRegressor & n\_estimators & 2, 5, 10, 20, 50 \\
& max\_depth & 1, 3, 6 \\
& min\_child\_weight & 1, 3, 5 \\
& gamma & 0, 0.5, 1\\
& subsample & 0.8, 1 \\
& colsample\_bytree & 0.2, 0.4, 0.6, 0.8 \\
& eta & 0.05, 0.1, 0.3, 0.5, 0.7, 0.9 \\
& lambda & 0, 1, 2, 5, 10 \\
SelectKBest$^2$ & k & 50, 100, 250, 500, 1000, 2500, 5000, 10000, 15000 \\
\hline
\end{tabular}
\end{table}
\small
\noindent $^1$ The number of trees in the Random Forest was fixed to 400 and therefore was not treated as a hyperparameter.

\noindent  $^2$ SelectKBest was only used in additional analyses using feature vectors extracted from the original VGG16 network without fine-tuning to account for the very high dimensionality.

\newpage
\begin{table}[!h]
\caption{Accuracy in the test data of the fine-tuned VGG16 models predicting binary SEP and used for feature vector extraction. Note that one model per image type was trained. Models are ordered from the highest to the lowest average accuracy across the three SEP measures.}
\centering
\begin{tabular}{|c|ccc|}
\hline
Image type & Assets & Expenditure & Income \\
\hline
Light source & 0.77 & 0.70 & 0.64 \\
Latrine & 0.79 & 0.68 & 0.55 \\
Kitchen & 0.69 & 0.64 & 0.66 \\
Bathroom & 0.76 & 0.66 & 0.56 \\
Water source & 0.71 & 0.65 & 0.59 \\
Front door & 0.75 & 0.65 & 0.55 \\
Satellite (25m buffer) & 0.69 & 0.60 & 0.63 \\
Stove & 0.71 & 0.63 & 0.56 \\
Wall & 0.69 & 0.59 & 0.59 \\
Floor & 0.66 & 0.59 & 0.59 \\
Roof & 0.65 & 0.58 & 0.59 \\
Satellite (100m buffer) & 0.62 & 0.62 & 0.55 \\
Street view & 0.63 & 0.57 & 0.55 \\
\hline
\end{tabular}
\end{table}

\newpage
\begin{table}[!h]
\caption{Prediction accuracy in terms of Pearson ($r$) and Spearman ($\rho$) correlation and RMSE in the test data by SEP measure, model type, and algorithm in models using the original VGG16 CNN without fine-tuning for feature extraction.}
\centering
\begin{tabular}{|l|ccc|ccc|ccc|}
\hline
Model &  & Assets & & & Expenditure & & & Income & \\
 & $r$ & $\rho$ & RMSE & $r$ & $\rho$ & RMSE & $r$ & $\rho$ & RMSE \\
\hline
Satellite: & & & & & & & & & \\
\hspace{0.5cm}ElasticNet & 0.39 & 0.37 & 0.49 & 0.30 & 0.28 & 5,731.24 & 0.16 & 0.21 & 12,263.25 \\
\hspace{0.5cm}Random Forest & 0.51 & 0.47 & 0.45 & 0.28 & 0.28 & 5,768.68 & 0.20 & 0.27 & 12,177.29 \\
\hspace{0.5cm}XGBoost & 0.51 & 0.48 & 0.45 & 0.29 & 0.29 & 5,714.68 & 0.16 & 0.22 & 12,339.61 \\
Outdoor: & & & & & & & & & \\
\hspace{0.5cm}ElasticNet & 0.60 & 0.61 & 0.44 & 0.42 & 0.43 & 5,412.28 & 0.30 & 0.32 & 11,861.71 \\
\hspace{0.5cm}Random Forest  & 0.70 & 0.70 & 0.38 & 0.45 & 0.44 & 5,298.78 & 0.28 & 0.33 & 11,977.49 \\
\hspace{0.5cm}XGBoost & 0.70 & 0.70 & 0.38 & 0.42 & 0.44 & 5,409.53 & 0.25 & 0.38 & 12,058.84 \\
Complete: & & & & & & & & & \\
\hspace{0.5cm}ElasticNet & 0.75 & 0.75 & 0.37 & 0.60 & 0.52 & 4,753.20 & 0.43 & 0.29 & 11,324.20 \\
\hspace{0.5cm}Random Forest & 0.84 & 0.84 & 0.30 & 0.61 & 0.55 & 4,779.78 & 0.43 & 0.45 & 11,508.91 \\
\hspace{0.5cm}XGBoost & 0.84 & 0.83 & 0.30 & 0.58 & 0.51 & 4,882.57 & 0.38 & 0.47 & 11,667.06 \\
\hline
\end{tabular}
\end{table}

\newpage
\begin{figure}[!h]
\caption{Map of the study area. The \textit{Manhiça sede} administrative unit is located within the Manhiça district, in southern Mozambique. Urban areas were extracted from the Copernicus Global 100 m land cover product while road data was downloaded from OpenStreetMap.}
\centering \unskip
\includegraphics[width=0.85\textwidth]{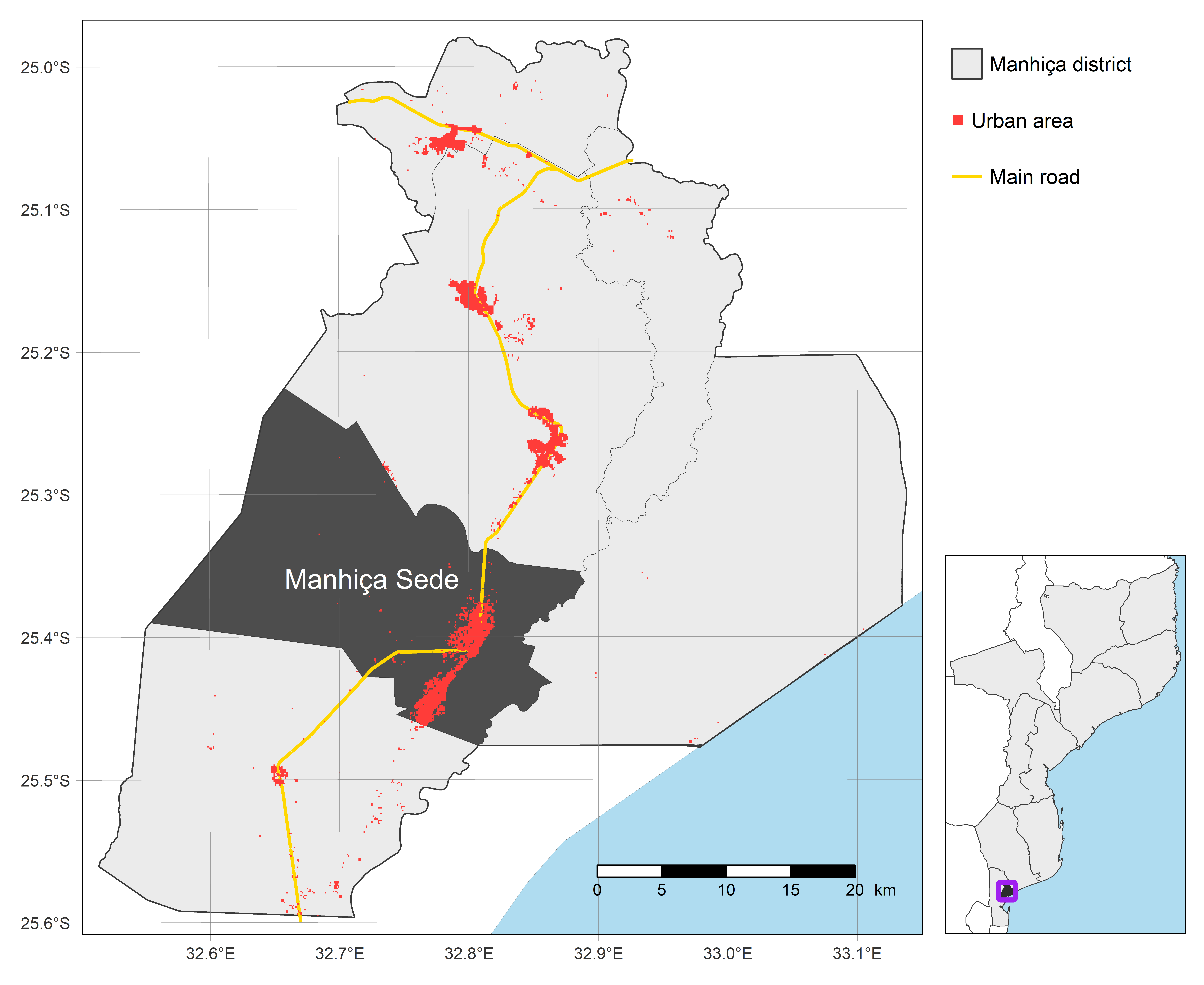}
\end{figure}   
\unskip

\newpage
\begin{figure}[!h]
\caption{Column principal coordinates of the first dimension of the MCA used to create the asset-based SEP measure.}
\centering \unskip
\includegraphics[width=0.7\textwidth]{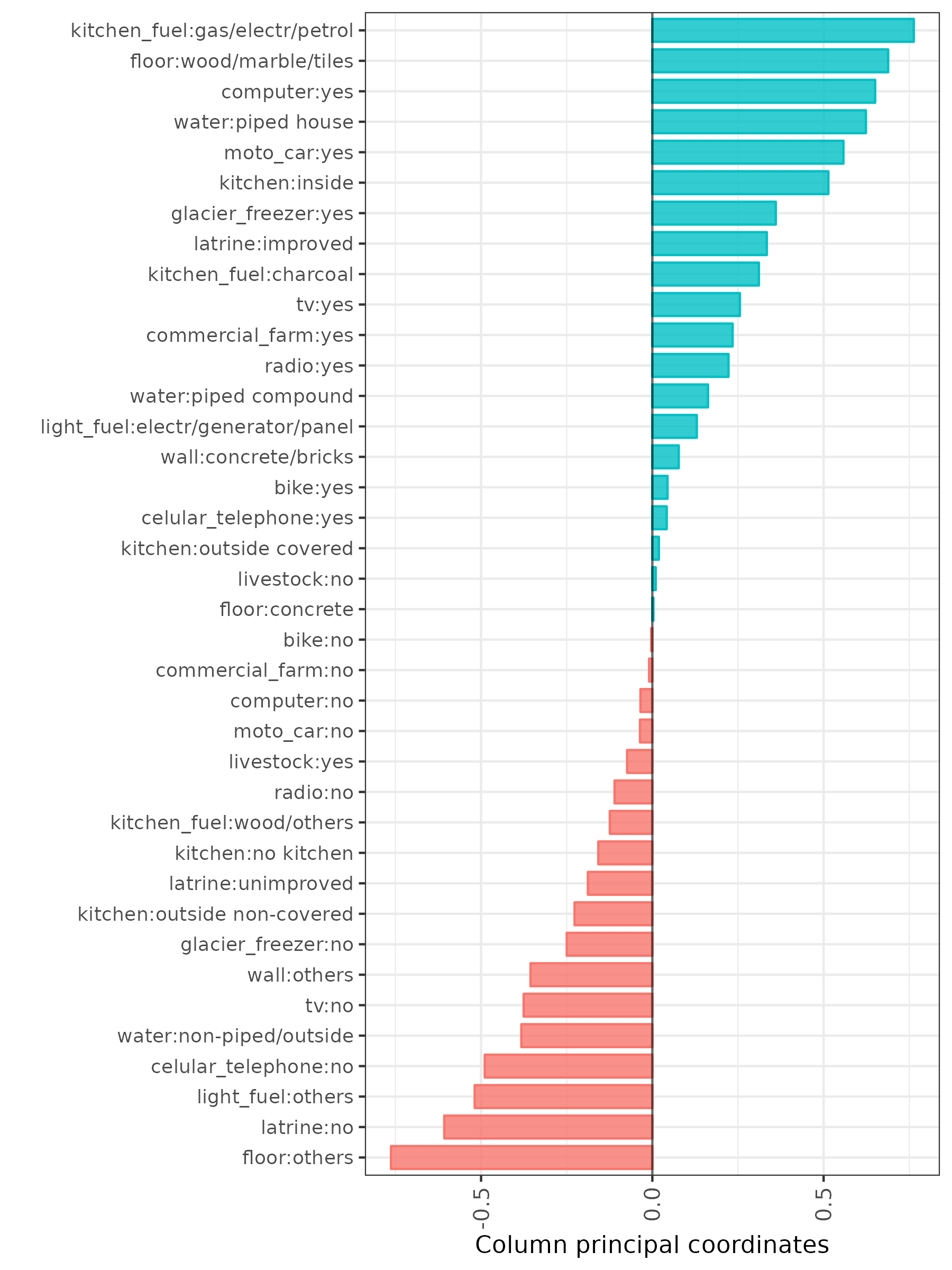}
\end{figure}   
\unskip

\newpage
\begin{figure}[!h]
\caption{Distribution of the assets, expenditure, and income-based SEP measures in train vs. test data as kernel density plots.}
\centering 
\includegraphics[width=1\textwidth]{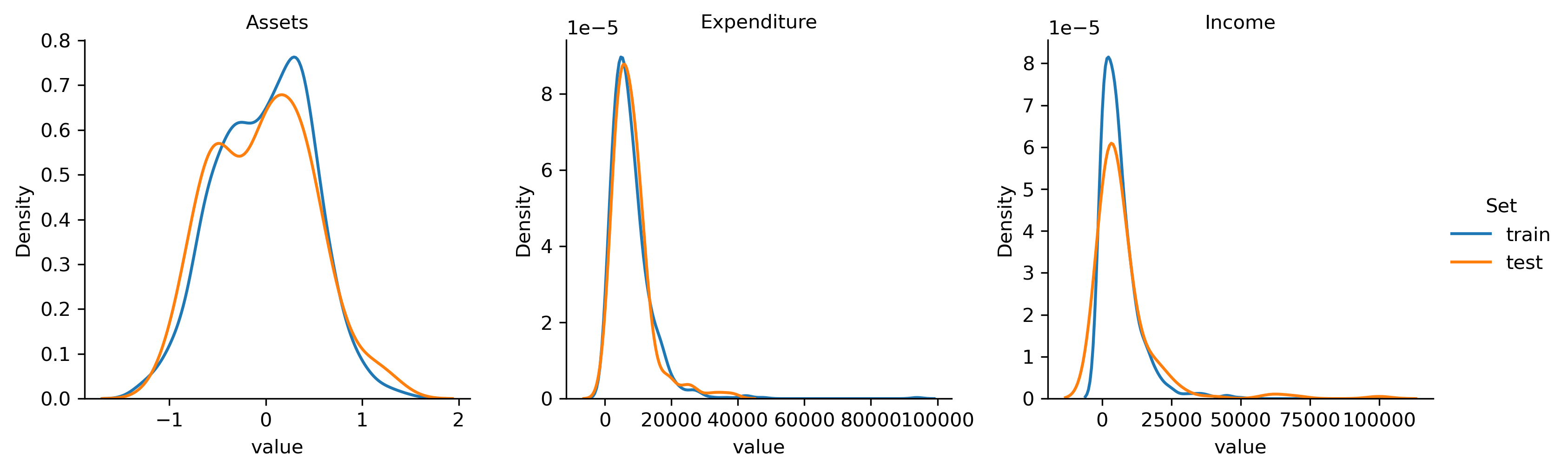}
\end{figure}   
\unskip

\newpage
\begin{figure}[!h]
\caption{Example images according to quartiles of the SEP measures (Q1 corresponds to households with lowest SEP measures and Q4 to the highest). Photographs for the wall, front door, kitchen, and bathroom image types are not included due to the potential risk of household identification.}
\centering 
\includegraphics[width=1\textwidth]{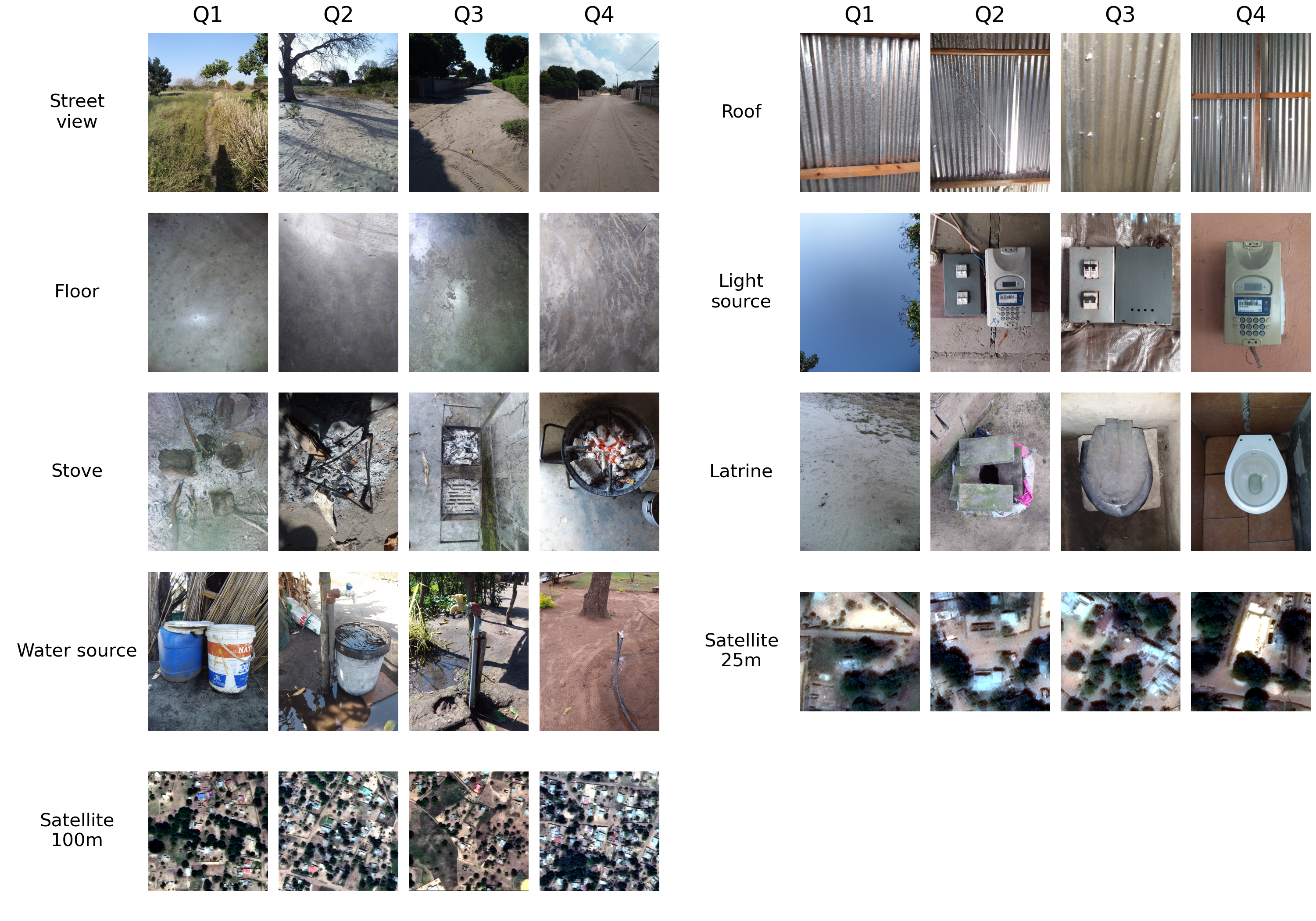}
\end{figure}   
\unskip

\newpage
\begin{figure}[!h]
\caption{Scatterplot of observed versus predicted asset-based SEP values in test data by algorithm type (rows) and predictor set (columns).}
\centering 
\includegraphics[width=0.9\textwidth]{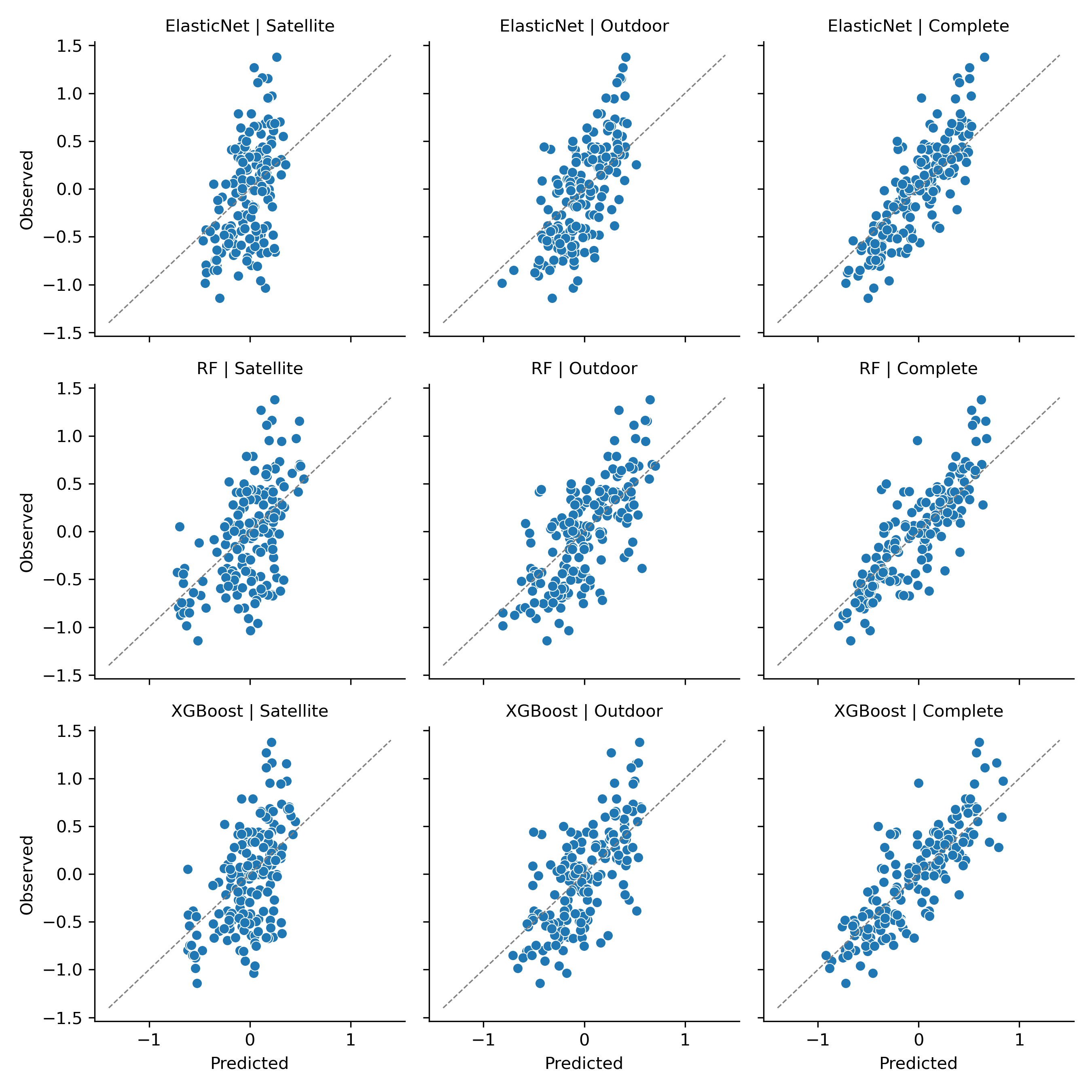}
\end{figure}   
\unskip

\newpage
\begin{figure}[!h]
\caption{Scatterplot of observed versus predicted expenditure-based SEP values in test data by algorithm type (rows) and predictor set (columns).}
\centering 
\includegraphics[width=0.9\textwidth]{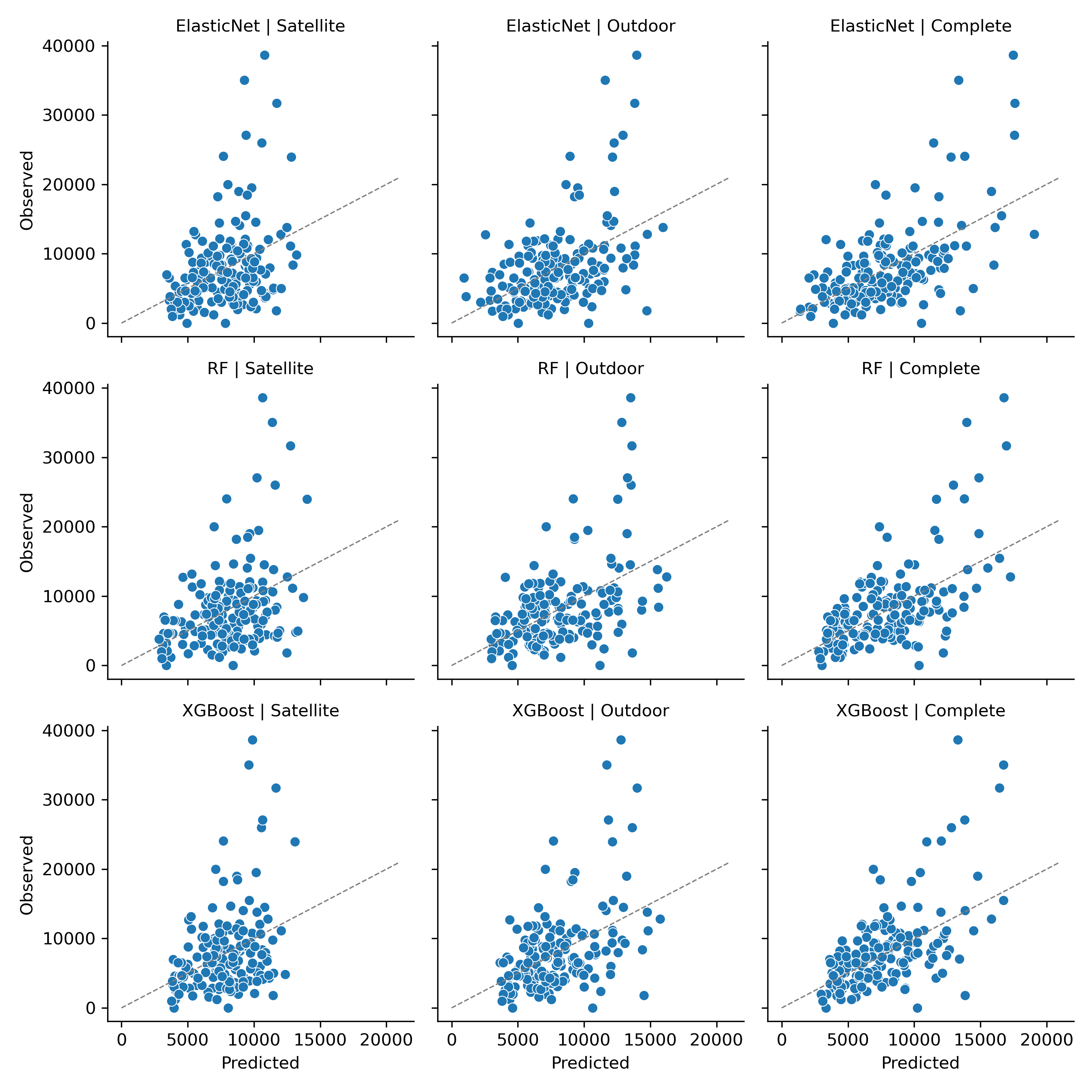}
\end{figure}   
\unskip

\newpage
\begin{figure}[!h]
\caption{Scatterplot of observed versus predicted income-based SEP values in test data by algorithm type (rows) and predictor set (columns).}
\centering 
\includegraphics[width=0.9\textwidth]{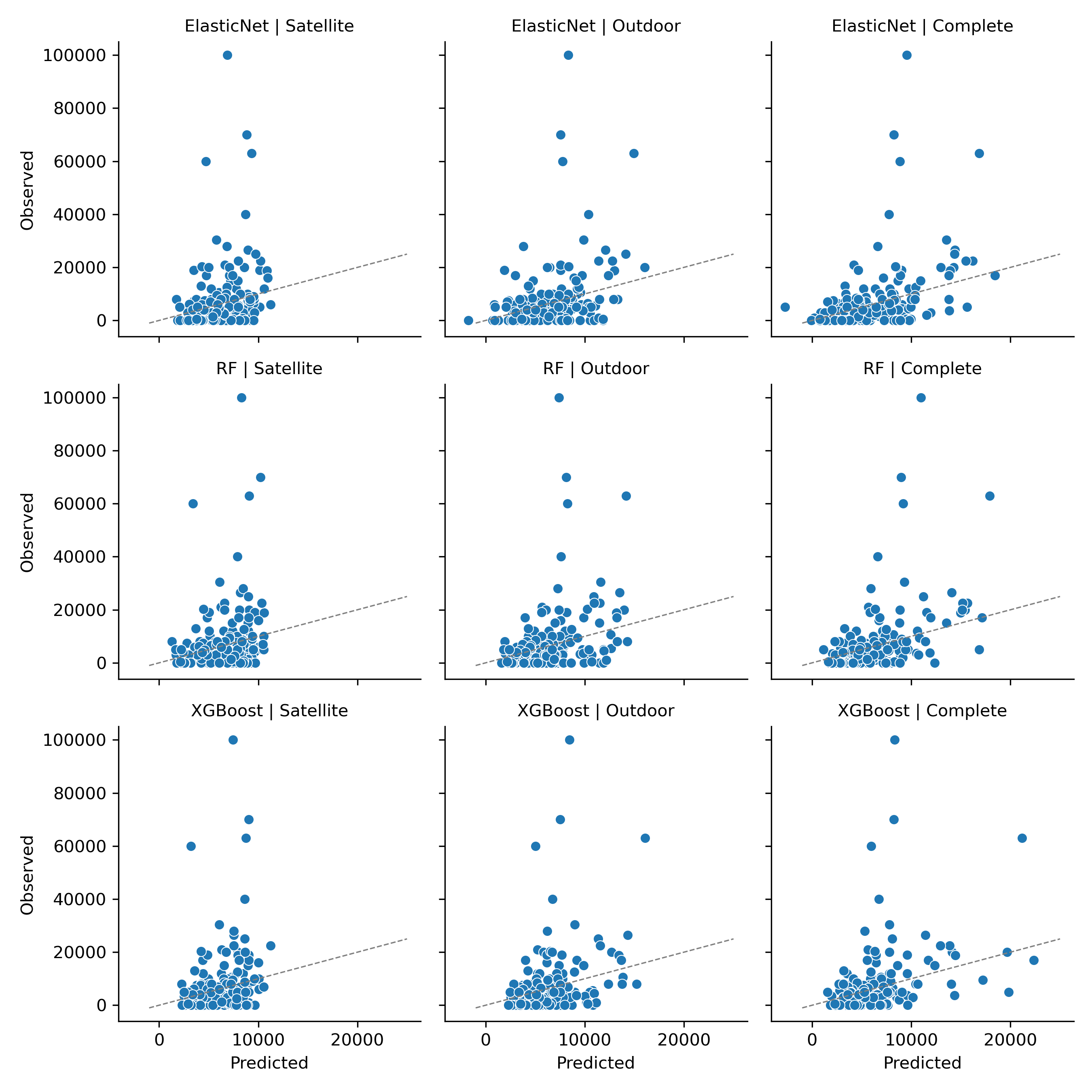}
\end{figure}   
\unskip

\newpage
\begin{figure}[ht!]
\caption{Street view images with the largest positive (top row) and negative (bottom row) SHAP values in random forest complete models. Images were selected according to the top and bottom average SHAP value ranks across the three SEP measures.}
\centering 
\includegraphics[width=0.85\textwidth]{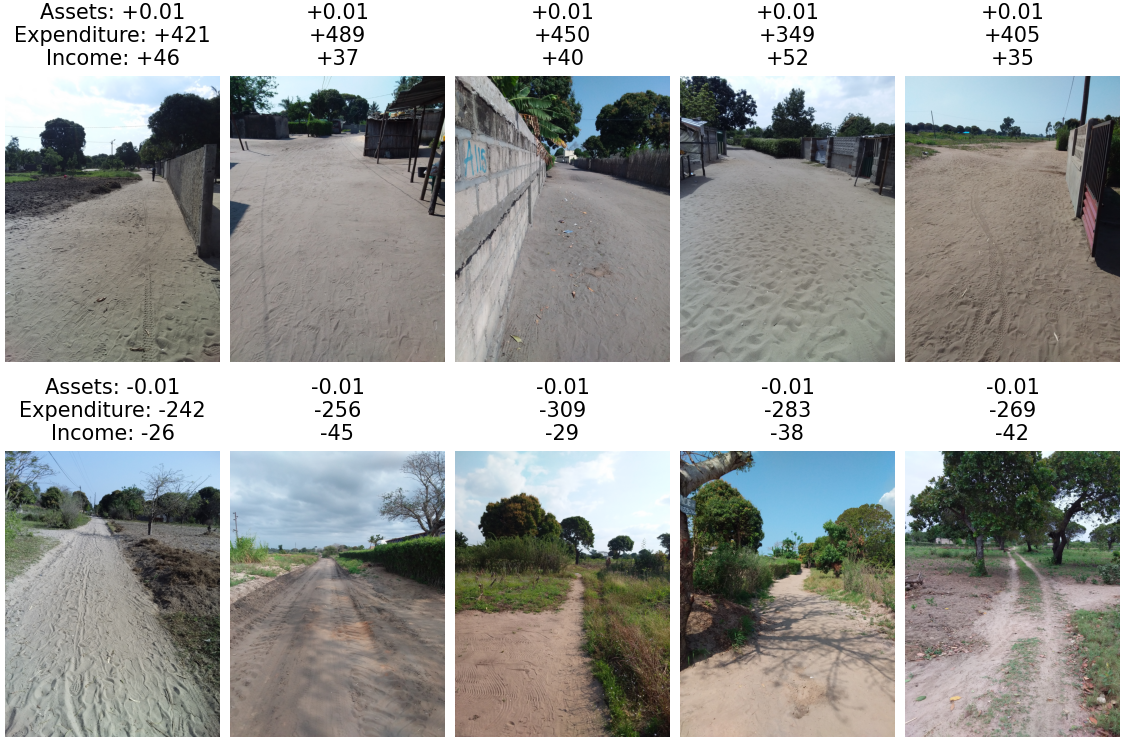}
\end{figure}   

\newpage
\begin{figure}[ht!]
\caption{Roof images with the largest positive (top row) and negative (bottom row) SHAP values in random forest complete models. Images were selected according to the top and bottom average SHAP value ranks across the three SEP measures.}
\centering 
\includegraphics[width=0.85\textwidth]{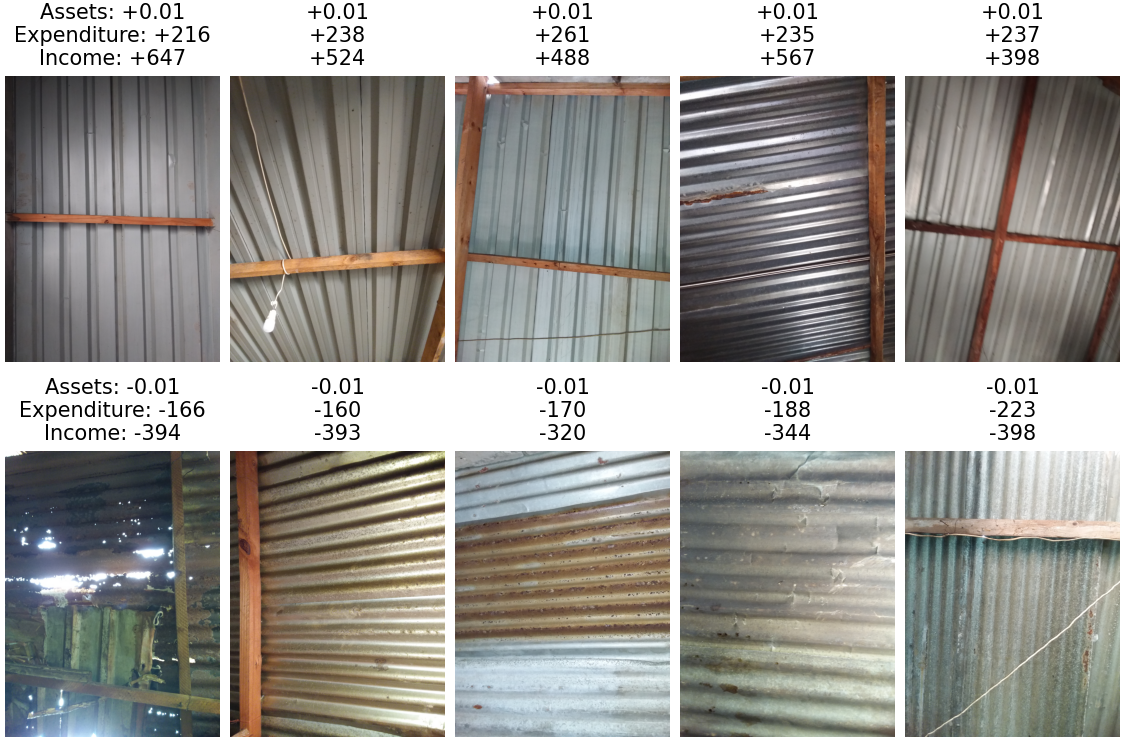}
\end{figure}   

\newpage
\begin{figure}[ht!]
\caption{Floor images with the largest positive (top row) and negative (bottom row) SHAP values in random forest complete models. Images were selected according to the top and bottom average SHAP value ranks across the three SEP measures.}
\centering 
\includegraphics[width=0.85\textwidth]{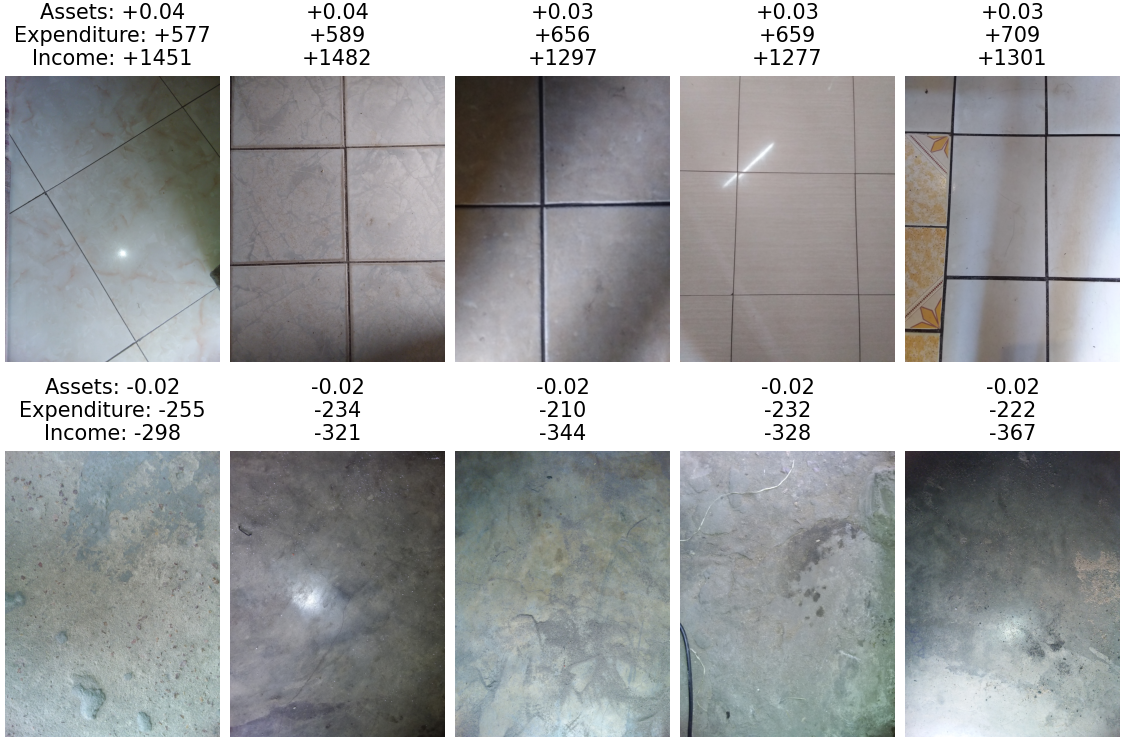}
\end{figure}   

\newpage
\begin{figure}[ht!]
\caption{Stove images with the largest positive (top row) and negative (bottom row) SHAP values in random forest complete models. Images were selected according to the top and bottom average SHAP value ranks across the three SEP measures.}
\centering 
\includegraphics[width=0.85\textwidth]{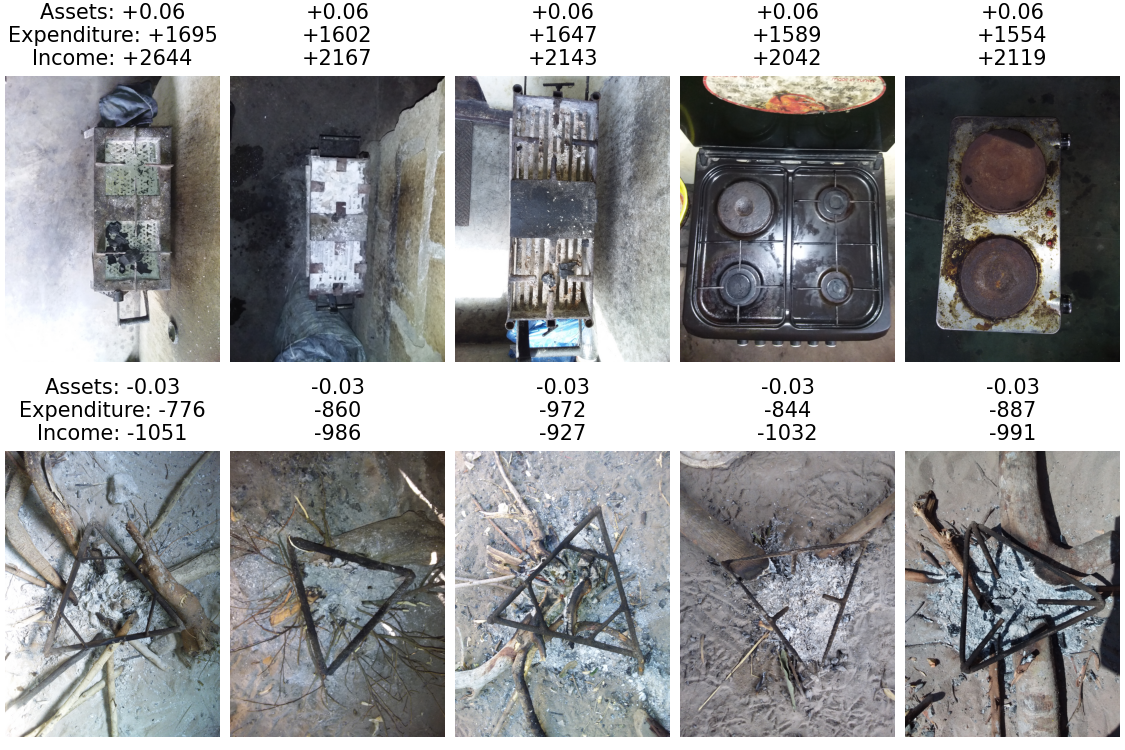}
\end{figure}   

\newpage
\begin{figure}[ht!]
\caption{Latrine images with the largest positive (top row) and negative (bottom row) SHAP values in random forest complete models. Images were selected according to the top and bottom average SHAP value ranks across the three SEP measures.}
\centering 
\includegraphics[width=0.85\textwidth]{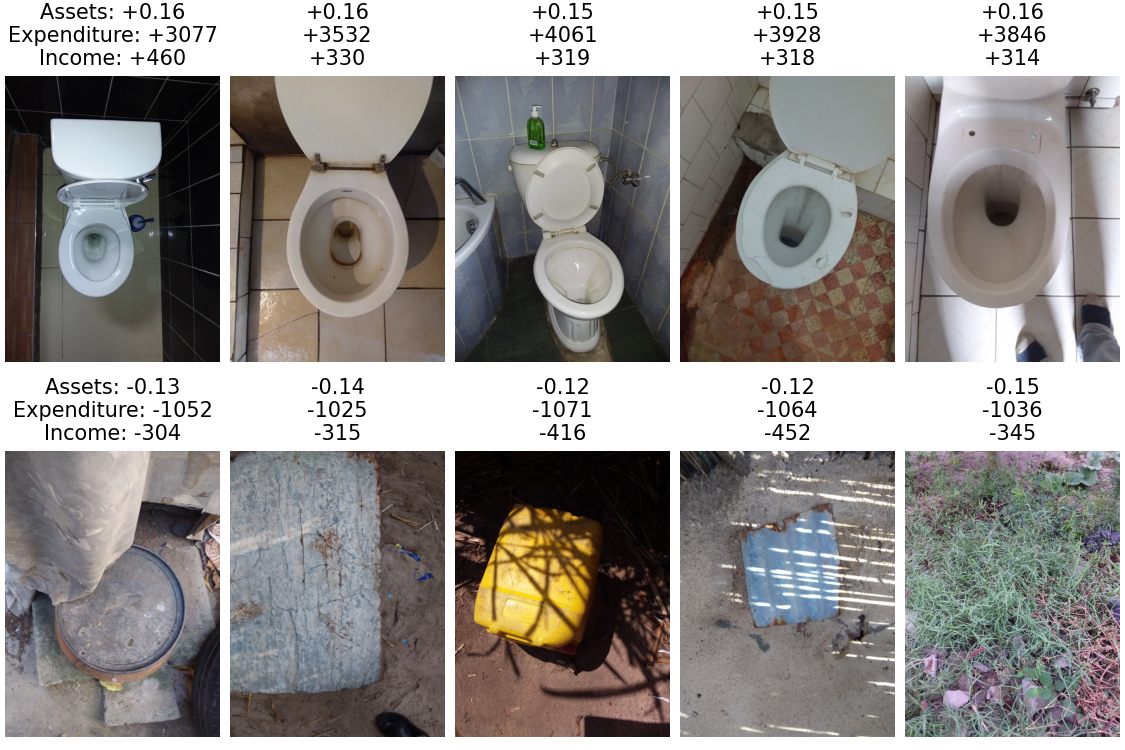}
\end{figure}   

\newpage
\begin{figure}[ht!]
\caption{Water source images with the largest positive (top row) and negative (bottom row) SHAP values in random forest complete models. Images were selected according to the top and bottom average SHAP value ranks across the three SEP measures.}
\centering 
\includegraphics[width=0.85\textwidth]{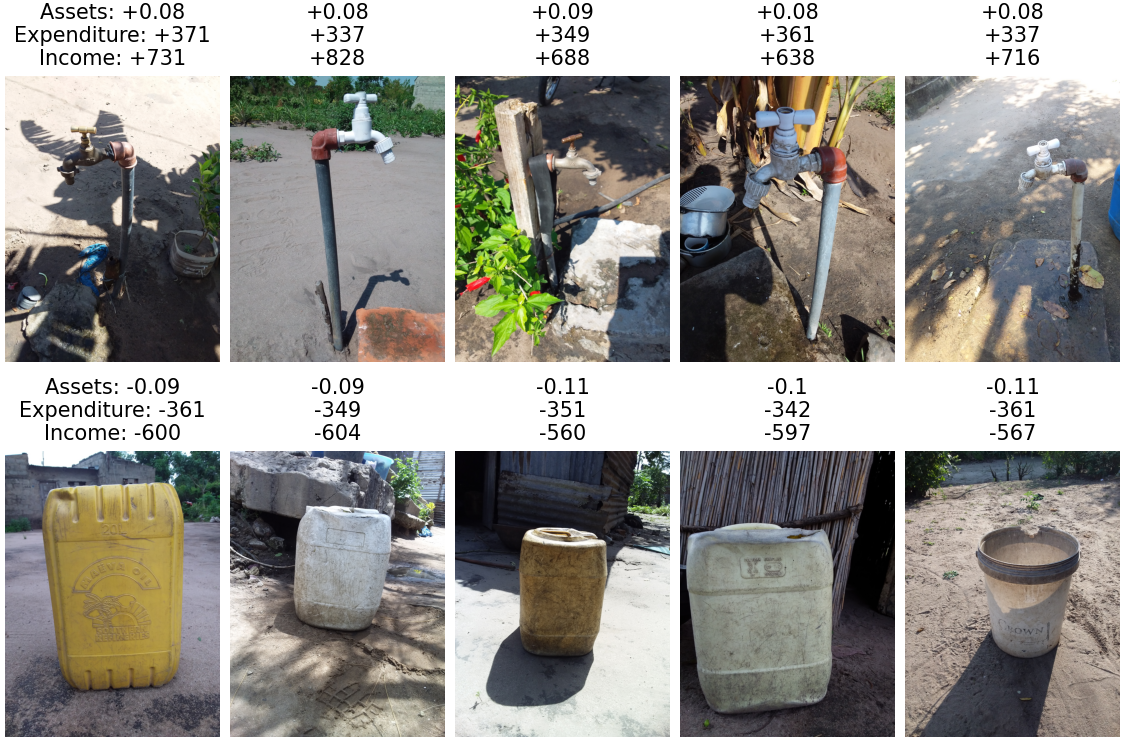}
\end{figure}   

\newpage
\begin{figure}[ht!]
\caption{Satellite (25m buffer) images with the largest positive (top row) and negative (bottom row) SHAP values in random forest complete models. Images were selected according to the top and bottom average SHAP value ranks across the three SEP measures.}
\centering 
\includegraphics[width=0.85\textwidth]{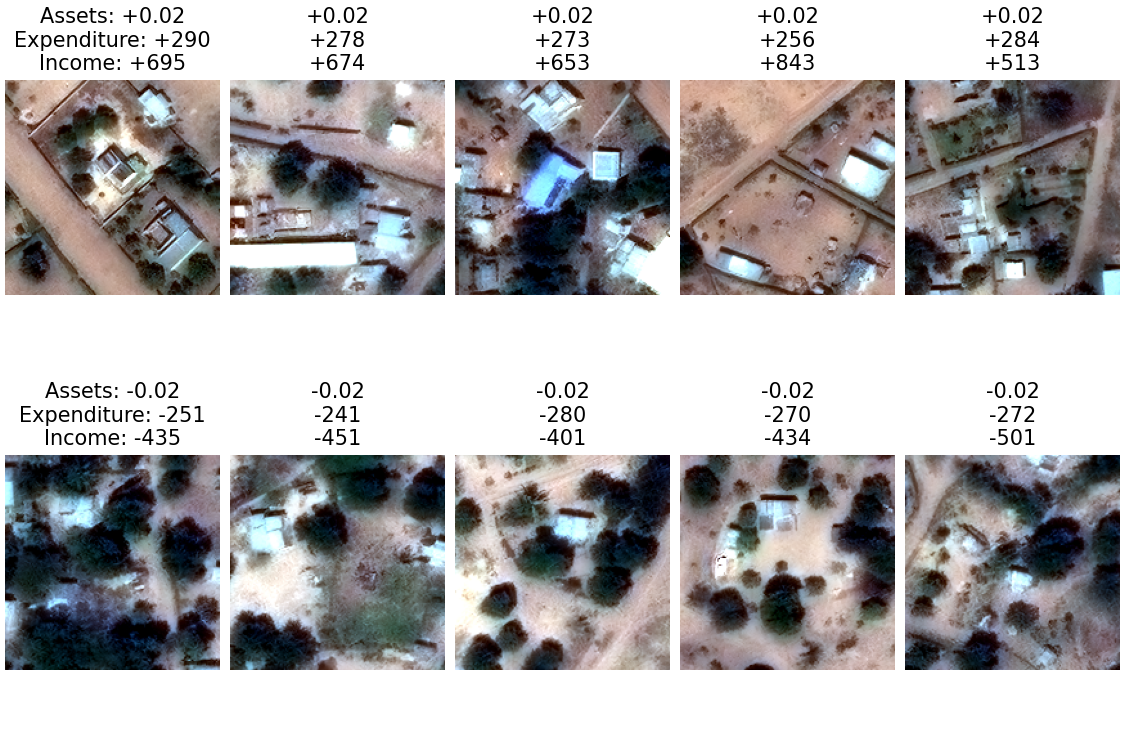}
\end{figure}   

\newpage
\begin{figure}[ht!]
\caption{Satellite (100m buffer) images with the largest positive (top row) and negative (bottom row) SHAP values in random forest complete models. Images were selected according to the top and bottom average SHAP value ranks across the three SEP measures.}
\centering 
\includegraphics[width=0.85\textwidth]{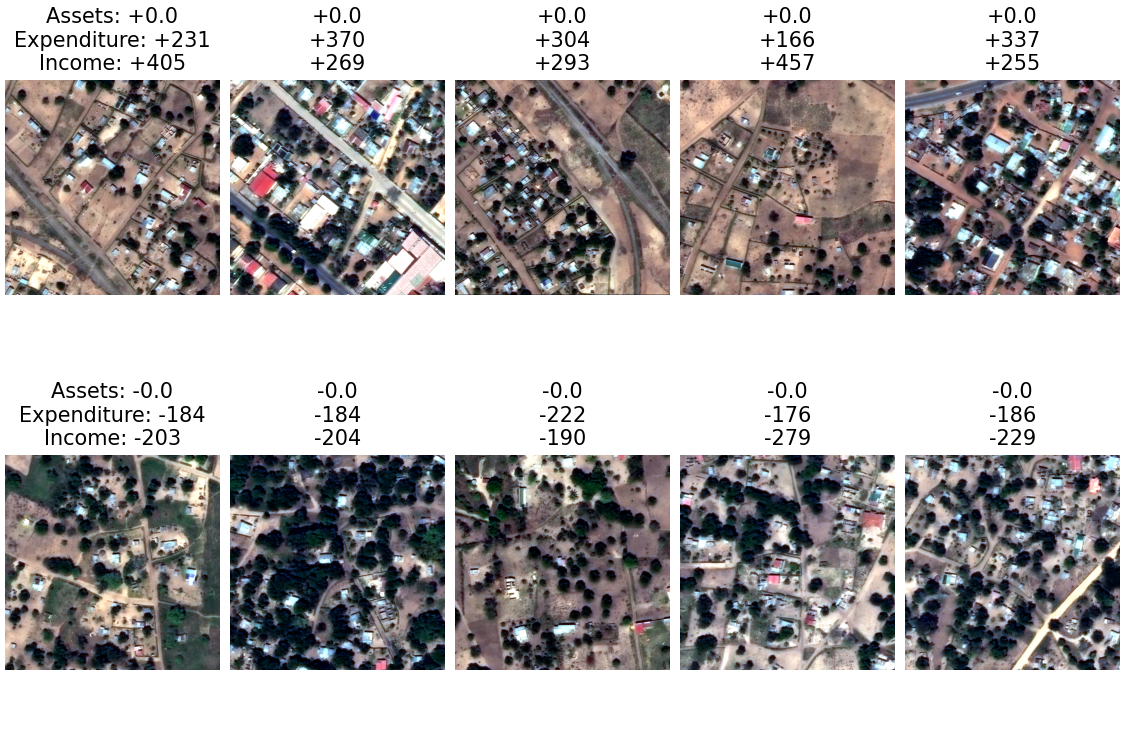}
\end{figure}